\documentclass[pmlr]{jmlr} 

\usepackage{booktabs}

\usepackage[switch]{lineno}

\newcommand{\equal}[1]{{\hypersetup{linkcolor=black}\thanks{#1}}}

\theorembodyfont{\upshape}
\theoremheaderfont{\scshape}
\theorempostheader{:}
\theoremsep{\newline}

\usepackage{hyperref}
\usepackage{xurl}

\usepackage{latexsym}
\usepackage{multirow}
\usepackage{graphicx}
\usepackage{bigstrut}
\usepackage{amsmath}
\usepackage{booktabs}
\usepackage{multirow}
\usepackage{tikz}
\usepackage{enumitem}
\usepackage{array}
\usepackage{mathtools}

\newcolumntype{C}[1]{>{\centering\arraybackslash}p{#1}}

\def \vec#1{{\mathbf{#1}}}

\newcommand{\bi}{\begin{itemize}}
\newcommand{\ei}{\end{itemize}}

\newcommand{\Lp}[1]{L^p (\Omega)}

\newcommand{\Linfinity}[1]{L^{\infty}(\Omega)}

\makeatletter
\def\hlinewd#1{%
\noalign{\ifnum0=`}\fi\hrule \@height #1 %
\futurelet\reserved@a\@xhline}
\makeatother

\jmlrvolume{}
\jmlryear{}
\jmlrworkshop{}
\jmlrpages{}

\title[Personalized FL on Clinical Datasets]{A Comprehensive View of Personalized Federated Learning on Heterogeneous Clinical Datasets}

 \author{%
 \Name{Fatemeh Tavakoli}\equal{These authors contributed equally} \Email{fatemeh.tavakoli@vectorinstitute.ai}\\
 \Name{D. B. Emerson}\footnotemark[1] \Email{david.emerson@vectorinstitute.ai}\\
 \Name{Sana Ayromlou} \Email{sana.ayromlou@vectorinstitute.ai}\\
 \Name{John Jewell} \Email{john.jewell@vectorinstitute.ai}\\
 \Name{Amrit Krishnan} \Email{amrit.krishnan@vectorinstitute.ai}\\
 \Name{Yuchong Zhang} \Email{yuchong.zhang@vectorinstitute.ai}\\
 \addr Vector Institute, Toronto, Canada
 \AND
 \Name{Amol Verma} \Email{amol.verma@mail.utoronto.ca}\\
 \Name{Fahad Razak} \Email{fahad.razak@mail.utoronto.ca}\\
 \addr Unity Health Toronto, Toronto, Canada
 }

\begin{document}

\maketitle

\begin{abstract}

Federated learning (FL) is increasingly being recognized as a key approach to overcoming the data silos that so frequently obstruct the training and deployment of machine-learning models in clinical settings. This work contributes to a growing body of FL research specifically focused on clinical applications along three important directions. First, we expand the FLamby benchmark \citep{Terrail1} to include a comprehensive evaluation of personalized FL methods and demonstrate substantive performance improvements over the original results. Next, we advocate for a comprehensive checkpointing and evaluation framework for FL to reflect practical settings and provide multiple comparison baselines. To this end, an open-source library aimed at making FL experimentation simpler and more reproducible is released. Finally, we propose an important ablation of PerFCL \citep{Zhang1}. This ablation results in a natural extension of FENDA \citep{Kim2} to the FL setting. Experiments conducted on the FLamby benchmark and GEMINI datasets \citep{Verma1} show that the proposed approach is robust to heterogeneous clinical data and often outperforms existing global and personalized FL techniques, including PerFCL.

\end{abstract}

\section{Introduction}

It is well-established that the robustness and generalizability of machine-learning (ML) models typically grow with access to larger quantities of representative training data \citep{Sun1}. While important in general settings, these properties are even more critical in clinical applications. For certain clinical tasks, studies have even produced estimates of minimum dataset sizes required to train models with sufficient utility to be useful in practical deployment \citep{Hosseinzadeh1}. In practice, however, a large portion of health datasets are generated and held by multiple institutions. Centralizing the data is often discouraged, if not impossible, due to strong regulations governing health data \citep{Herrin1}. Moreover, data annotation tends to be expensive, often requiring experts, such as doctors. Thus, large, single-institution datasets are frequently unavailable.

Federated learning (FL), first introduced by \citet{McMahan1}, provides an avenue for training models in distributed data settings without requiring training data transfer. The focus of this work is \emph{cross-silo} FL \citep{Kairouz1}, where participants represent a small number of reliable institutions with sufficient computing resources. In particular, we focus on the challenging setting of distributed heterogeneous datasets. Such datasets occur in many FL settings and are particularly common in clinical environments, as the statistical properties of local datasets are influenced by, for example, disparate patient populations and varying physical equipment. Such datasets often require special treatment in FL frameworks to avoid issues with convergence. Personalized FL methods have proven promising for such non-identically and independently distributed (non-IID) datasets.

Another meaningful challenge in the study of FL methods in clinical settings is the lack of access to high-quality benchmarking pipelines and real-world medical datasets. Therefore, we provide a thorough benchmark of state-of-the-art personalized and non-personalized FL methods, including a promising ablation, evaluated with a comprehensive framework on real-world public and private clinical datasets. The main contributions are as follows.

\begin{itemize}

\item We advance existing work in personalized FL benchmarking to consider the challenging setting of heterogeneous clinical data. In so doing, marked improvements in the performance of many FL approaches are demonstrated for the datasets considered. This includes establishing that personalized methods offer significant advantages over non-personalized approaches for real-world clinical tasks.

\item We advocate for a more comprehensive evaluation framework for FL methods. This includes utilizing reasonable checkpointing strategies, bringing FL training into better alignment with standard ML practices, and evaluating model performance from several perspectives. With this work, an open-source library aimed at making FL research and evaluation easier and more reproducible is released.\footnote{All experimental code is available at: \url{https://github.com/vectorInstitute/FL4Health}}

\item A novel ablation of the recently introduced PerFCL algorithm \citep{Zhang1} is proposed that results in a natural extension of the frustratingly easy neural domain adaptation (FENDA) method \citep{Kim2} to the FL regime and, thus, is termed FENDA-FL. Such an ablation proves to be important as experiments demonstrate that the approach often outperforms existing FL techniques.

\end{itemize}

\subsection*{Generalizable Insights in the Context of Healthcare}

The results in this work provide important insights and value to those interested in leveraging FL in clinical applications. The experiments provide a thorough evaluation of carefully selected personal and non-personal FL methods, as discussed in Appendix \ref{pfl_method_choice}, on real-world clinical tasks, demonstrating that personalized methods provide marked performance improvements. Notably, in most settings, personalized FL approaches even outperform models trained on centralized data. The evaluation framework put forward herein provides rigorous assessment of the studied FL methods, and the proposed checkpointing approaches ensure optimal performance. Finally, the proposed ablation of PerFCL, FENDA-FL, produces consistently strong results across most of the clinical tasks investigated, see Table \ref{fl_method_rank_table}, making it a compelling candidate for further application in healthcare.

\section{Related Work}

\subsection{Standard and Personalized FL}
FL has been successfully applied in many contexts \citep{Hard1, Wang1, Dupuy1} to jointly learn ML models on distributed data. It is an active area of research along many dimensions, from communication efficiency \citep{Passban1, Isik1} to improved optimization \citep{McMahan3}. A central challenge in FL is overcoming heterogeneous data distributions arising from variations in the underlying data generating process of clients. To address this, personalized FL approaches aim to federally train models that produce elevated performance on each participating client's individual data distribution. Several state-of-the-art personalized FL techniques, FedPer \citep{Arivazhagan1}, Ditto \citep{Ditto1},  APFL \citep{Deng1}, and PerFCL \citep{Zhang1}, are extensively evaluated herein. The FENDA-FL approach, discussed in detail below, performs comparably or better in most settings.

\subsection{FL for Clinical Tasks and Benchmarking}
While the number of FL applications in clinical settings is growing \citep{Terrail2, Dayan1}, the approach is still relatively under-utilized, despite its potential benefits. This is due, in part, to a dearth of research and tooling focused on this important area. Some existing works focus on the utility of FL for specific clinical applications \citep{Andreux1, Andreux2, Chakravarty1, Wang5, Gunesli1}. The FLamby benchmark \citep{Terrail1} aims to standardize the datasets in these studies to facilitate systematic evaluation of FL in healthcare. While the benchmark has significant utility, the limited evaluation protocols and omission of personalized FL methods narrow the insights of the benchmark in the context of clinical applications. While there are some benchmarks focused on personalized FL methods \citep{Matsuda1, Chen1}, evaluation of personalized FL in the healthcare domain is unexplored. 

Our work aims to provide a broader account of the performance of FL, including personalized methods, on clinical datasets using reasonable checkpointing and evaluation strategies that are more closely aligned with standard ML practices. To this end, we leverage several datasets from the FLamby collection, improving upon many of the originally reported baselines and establishing new performance standards. We also extend the understanding of the practical use of FL in clinical settings by measuring results on datasets from the GEMINI consortium. The experiments demonstrate that personalized FL methods generally improve upon non-personalized FL approaches and highlight the importance of intelligent checkpointing strategies.

\section{Methodology} \label{methodology}

In this work, the performance and utility of non-personal and personal FL algorithms are measured. Non-personalized methods are important when a single model is desired across clients and can be effective, even in heterogeneous data settings. On the other hand, personalized FL techniques produce models that benefit from cross-client aggregation, while retaining client-specific representations. As such, these approaches provide a promising balance between global and local information and are of specific interest in clinical settings, where performance is of the utmost importance. 

As in \citet{Terrail1}, the experiments to follow incorporate several foundational FL techniques as baselines along with more recent approaches. The methods evaluated are FedAvg \citep{McMahan1}, FedAdam \citep{McMahan3}, FedProx \citep{Tian1}, SCAFFOLD \citep{Karimireddy1}, MOON \citep{Li2}, FedPer \citep{Arivazhagan1}, Ditto \citep{Ditto1}, APFL \citep{Deng1}, and PerFCL \citep{Zhang1}. The choice of personalized methods studied in this work is discussed in Appendix \ref{pfl_method_choice}. Two non-FL training setups are also considered. The first, central training, pools all data to train a single model. The second is local training, where each client trains a model solely on local data, resulting in a model for each client with no global information.

\subsection{Non-Personalized FL Methods}

The original FedAvg algorithm, which applies weighted parameter averaging, remains a strong FL baseline. FedAdam is a recent extension of FedAvg that incorporates server-side first- and second-order momentum estimates and has shown some robustness to data drift. The FedProx and SCAFFOLD methods aim to address issues associated with non-IID datasets by correcting local-weight drift through a regularization term or control variates, respectively. MOON builds on these ideas, applying a contrastive loss to constrain drift in the client models' feature representations. Each method trains a global set of weights, updated at each server round, shared by all clients.

\subsection{Personalized FL Methods}

For the personalized FL baselines, each client possesses a unique model. FedPer splits the models into a feature extractor and classifier. The feature extractor is shared among clients and federally trained, while the classifier is unique to each client. Ditto trains a global model with FedAvg and distinct personal models. During client-side training, local model optimization is constrained using a weighted $\ell^2$-penalty between the global and local weights. In APFL, predictions are made through a convex combination of twin models. One is federally trained using FedAvg; the other incorporates only local updates. The combination parameter, $\alpha$, is adapted during training using the gradients of the global and local models. Finally, PerFCL extends MOON to consider two feature extractors, one is shared by all clients and the other is specific to each client along with a local classification layer. The feature extractors are constrained using different contrastive losses.

\subsection{FENDA-FL} \label{fenda_introduction}

FedPer, APFL, and PerFCL fall under the category of parameter decoupling methods. Such methods reserve a subset of models weights for learning global information and others for local information. In the results below, this type of decoupling provides useful flexibility in clinical settings. For PerFCL, the authors consider a model with parallel global and local feature extractors feeding into a locally trained classifier. The latent spaces of the extractors are constrained with contrastive loss functions that attempt to keep each client's global latent space close to the aggregated global feature space and to push local features away from these spaces. These losses are weighted with hyper-parameters $\mu$ and $\gamma$. Experiments in the original paper show that PerFCL improves upon MOON and other FL baselines. However, therein, PerFCL is not compared against other personalized methods. Moreover, no ablation studies are considered to quantify the benefit associated with the architecture compared to the loss. As such, we study the effect of removing the contrastive losses.

Removal of the contrastive losses results in an approach that directly extends the FENDA method, originally designed for domain adaptation, to FL. In FENDA, each model associated with a domain has two feature extractors. One is trained on domain-specific data, while the other is shared and trained across all domains. The features are then combined and processed by a domain-specific classification head. The shared feature extractor is free to learn robust, domain-agnostic features, while the domain-specific extractor learns features important in making accurate local predictions.

Assume there are $M$ clients, with local datasets $D_{i} \subset X \times Y$, for $i \in \{1, \dots, M\}$. As with PerFCL, each FENDA-FL client model consists of a global feature extractor, $f_{\vec{w}}: X \rightarrow \mathbb{R}^{H_G}$, a local feature extractor $f_{\vec{w}_i}: X \rightarrow \mathbb{R}^{H_i}$, and a local classification head $g_{\theta_i}: \mathbb{R}^{(H_G + H_i)} \rightarrow Y$. For a given input sample $\vec{x} \sim P_i(X)$, $f$ and $f_i$ map $\vec{x}$ to representations $\vec{z}$ and $\vec{z}_i$ which are then concatenated and passed into $g_{\theta_i}$ to yield a prediction, $\hat{y}_i$, as
\begin{align*}
    \hat{y}_i = g_{\theta_i} \left(
    \begin{bmatrix}
        \vec{z}  \\
        \vec{z}_i  \\
  \end{bmatrix}\right) = g_{\theta_i} \left(
    \begin{bmatrix}
        f_{\vec{w}}(\vec{x})  \\
        f_{\vec{w}_i}(\vec{x})  \\
  \end{bmatrix} \right).
\end{align*}
The global extractor weights, $\vec{w}$, are aggregated across clients via FedAvg at each round. The local extractor and classification head weights, $\vec{w}_i$ and $\theta_i$, are exclusively learned through SGD, or its variants, on each client. Appendix \ref{formal_fenda_desription} discusses distribution drift robustness and FENDA-FL's relationship to domain adaptation. As formulated, FENDA-FL is not limited to clinical domains. Nonetheless, in this work, we are primarily interested in the performance of FL approaches for heterogeneous clinical applications.

FENDA-FL has several advantages over PerFCL, beyond a simpler setup. Removal of the contrastive losses eliminates the need to tune two hyper-parameters. Further, it has been observed that similar loss-induced constraints may hinder learning, especially when hyper-parameters are poorly calibrated \citep{Tian1}. Without the contrastive losses, FENDA-FL is also a more flexible approach - global and local modules need not have the same latent space, allowing for the injection of inductive bias through architecture variation. As shown in Appendix \ref{architecture_ablation}, this is useful in clinical settings to produce optimal performance.

\subsection{Federated Checkpointing and Evaluation} \label{checkpoint_and_eval}

Federated evaluation is a nuanced and under-explored process, yet it is a critical component, especially in clinical pipelines \citep{Karargyris2021FederatedBO, Matsuda1}. We propose a straightforward checkpointing and evaluation framework for FL models, building on the foundations in \citep{Terrail1} and better aligning with standard ML model practices. In many works \citep{Tian1, Karimireddy1, Passban1}, FL training proceeds for a fixed number of server rounds. Thereafter, the final model is evaluated on held-out data, representing either a central or distributed test set. However, as with standard ML training, the model produced at the end of a training trajectory is not always optimal due to phenomena such as over-fitting. 

We propose splitting each clients' dataset into train and validation sets. In the experiments, the split used is 80--20. The validation splits enable two distinct checkpointing approaches. After each round of local training and aggregation, the clients evaluate the model on their local validation sets. In \emph{local checkpointing}, each client uses their local validation loss to determine whether a new checkpoint should be stored. Alternatively, in \emph{global checkpointing}, the client losses are aggregated by the server using a weighted average. This average is used to checkpoint a global model shared by all clients. In both cases, the best model is not necessarily the one produced by the final server round. In the local case, clients are free to keep the model they deem best. In the global strategy, all clients share the same model from some point in the server rounds. Personalized FL strategies do not have a single global model and, therefore, only use local checkpointing. When local checkpointing is used, each client loads its respective model to be evaluated on that client's test data.

\paragraph{Baselines} To evaluate checkpointed models, we advocate for at least three separate baselines to be considered, two of which are part of the FLamby benchmarks. Models trained on \emph{central} data are a common gold standard for FL performance. In the experiments, a single model is trained on data pooled from each of the clients and checkpointed based on a validation loss. The model is evaluated on each client's test set and the performance uniformly averaged. In \emph{local} training, a client trains a model on its own training data. The model is shared with all clients and evaluated on their test data, including that of the original client. This setting mimics the scenario where, for example, a hospital distributes a model to other hospitals. Again, the measured metrics are averaged across clients. An important alternative setting, referred to here as \emph{siloed}, is one where each client trains its own model and evaluates on its local test data. The results are then averaged across the clients. This simulates a strong baseline where, for example, the problem is important enough that each hospital has collected enough data to train a model. If an FL model does not produce better performance, there is significantly less incentive to participate in collaborative training.

\paragraph{Reproducibility} 

To quantify stability and smooth randomness in training, five FL training runs with distinct validation splits are performed for each method. Within each run, the performance across clients is averaged. The mean and $95$\% confidence intervals for this average are reported. As mentioned above, code to reproduce the experiments is released with this work along with an open-source library.

\begin{figure}[ht!]
    \centering
    \subfigure[]{
    \includegraphics[width=0.49\linewidth]{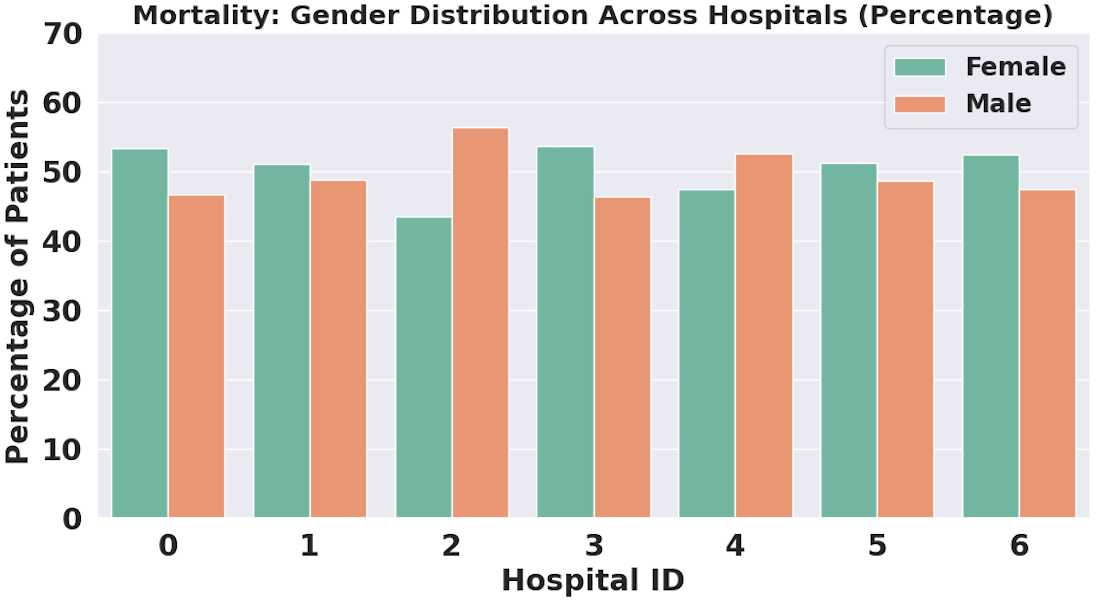}}
    \subfigure[]{\includegraphics[width=0.49\linewidth]{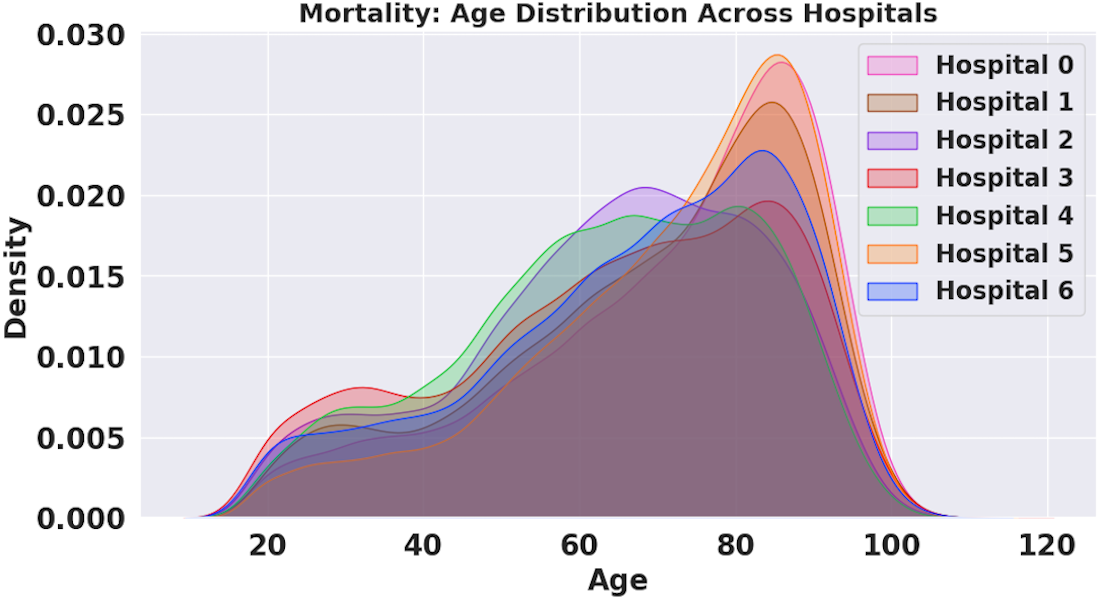}} \\
    \subfigure[]{\includegraphics[width=0.49\linewidth]{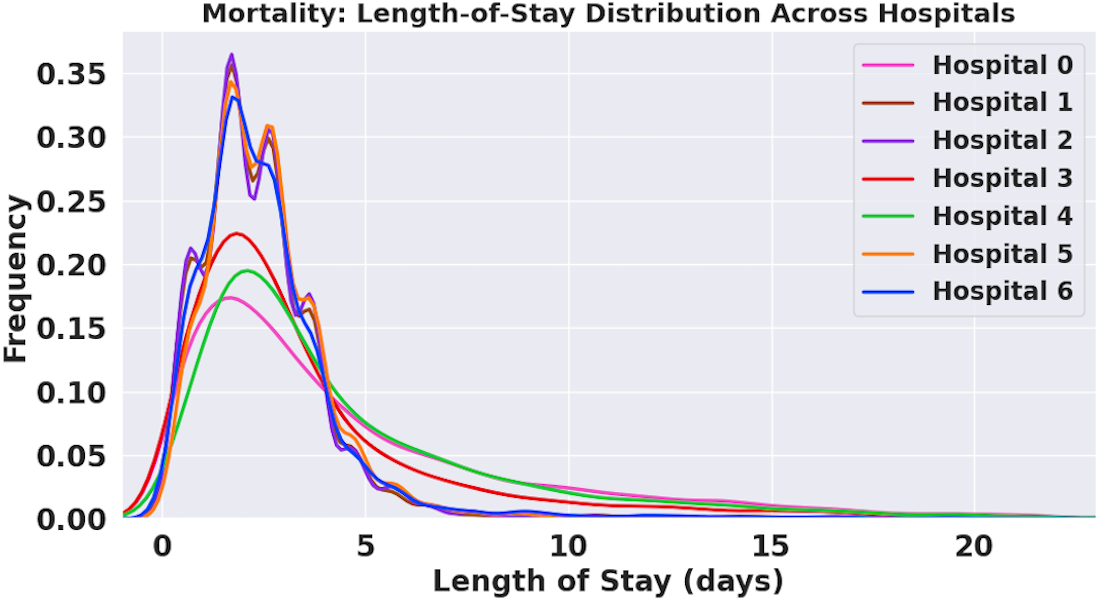}}
    \subfigure[]{\includegraphics[width=0.49\linewidth]{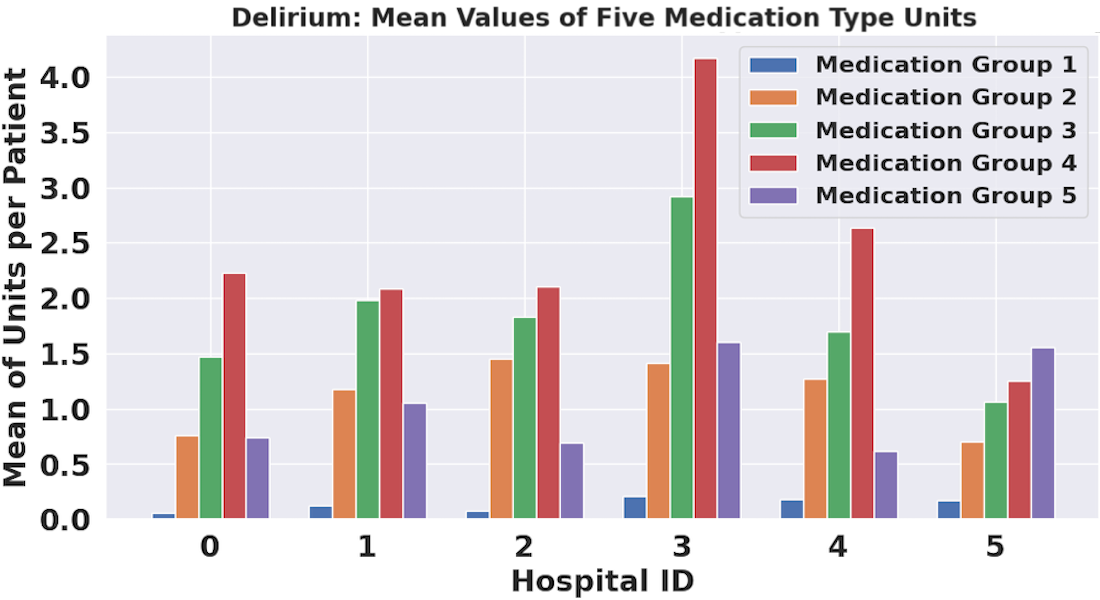}} \\
    \subfigure[]{\includegraphics[width=0.49\linewidth]{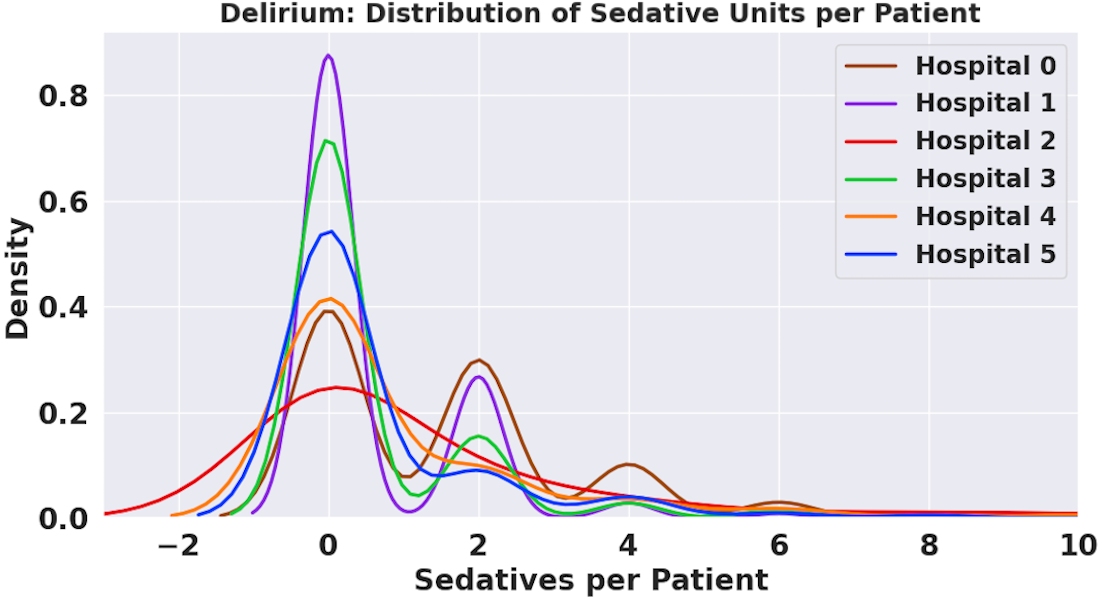}}
    \subfigure[]{\includegraphics[width=0.49\linewidth]{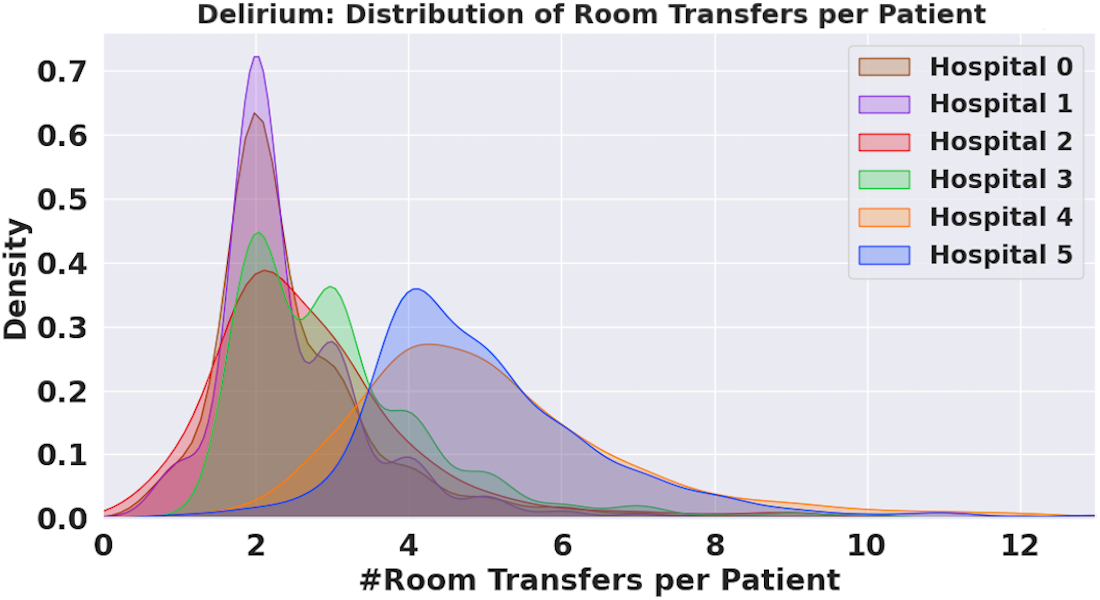}}
    \caption{Feature distributions for patient gender (a), age (b), and length of stay at care units (c) for the Mortality dataset and  units per medication group (d), sedative units (e), and number of room transfers per patient (f) in the Delirium dataset. Note that some features are normalized and smoothed for visualization.}
    \label{gemini_distribution_figure}
\end{figure}

\section{Clinical Datasets} \label{clinical_datasets}

In this work, five distinct clinical datasets are considered. Three are drawn from the FLamby benchmark \citep{Terrail1} and two are privately held within the GEMINI consortium \citep{Verma1}. Each represents a clinically relevant task and data derived from real patients. The selected FLamby datasets are Fed-Heart-Disease which is a binary task predicting whether a patient has heart disease from a collection of measurements, Fed-IXI which considers MR images of patients' heads with the target of 3D segmentation of brain-tissue, and Fed-ISIC2019 which is a multi-class dermatology dataset consisting of 2D images. The task is to classify the images into one of eight different melanoma categories. More details on these datasets are found in Appendix \ref{flamby_data_hps}.

The GEMINI datasets consist of clinical data from seven hospitals in Canada. The data is anonymized by removing patient identifiers, yet still contains highly sensitive patient-level information. As such, some dataset details are provided in the sections below and Appendix \ref{gemini_data_hps}, while other information is withheld in accordance with the GEMINI data sharing and research ethics protocols. The GEMINI tasks are critical to improving patient care and are distinct from the FLamby tasks in several ways. Predictions are based on tabular data with a large number of features. Clients have access to a sizeable number of labeled data points. Finally, label imbalance in both settings increases training complexity. Due to the sensitive nature of the data and GEMINI’s closed environment, the datasets and code are not publicly released. However, they are available to researchers with access to GEMINI.\footnote{GEMINI access may be requested at \url{https://geminimedicine.ca/access-gemini-data/}}

\subsection{GEMINI: Mortality}

This task is in-hospital mortality prediction, which determines the probability of a patient's latest encounter ending in mortality. The cohort consists of patients admitted to general internal medicine (GIM) wards and discharged between 2015 to 2020, prior to COVID-19, at seven large hospitals. De-identified electronic health records (EHR) and administrative information from a total of $114,662$ patients are used for training, with $28,665$ data points reserved for testing across all hospitals. Data heterogeneity across hospitals for a selection of features is visualized in Figures \ref{gemini_distribution_figure}(a)--\ref{gemini_distribution_figure}(c). Additional details and patient characteristics for a selection of features are described in Appendix \ref{gemini_mortality_details}. Note that related heterogeneity visualizations for the FLamby datasets appear in the original paper.

\subsection{GEMINI: Delirium} \label{delirium_dataset}

The goal of this task is to determine whether delirium occurred during a hospital stay in order to facilitate quality measurement and benchmarking, because delirium is a leading cause of preventable harm in hospitals. Labeling delirium cases is labor-intensive, requiring specialized medical knowledge. Even in the GEMINI consortium, only a relatively small population has been labeled, see Table \ref{fl_dataset_distributions}. The total cohort consists of $5,770$ admissions including EHR and administrative features of patients admitted to GIM wards from 2010 to 2019 across six hospitals. Features such as lab measurements, medications, and diagnosis are used to perform binary classification. Figures \ref{gemini_distribution_figure}(d)--\ref{gemini_distribution_figure}(f), exhibit heterogeneity for a subset of features. More details are available in Appendix \ref{gemini_delirium_details}.

\section{Experiments and Results} \label{results}

In this section, results associated with three experiments are discussed. The main benchmarking and evaluation results for all datasets are detailed in Section \ref{main_results}. In Section \ref{checkpoint_ablation}, ablation studies are performed to provide a deeper analysis of the proposed checkpointing approaches. The third experiment, discussed in Section \ref{delirium_het_study}, considers the robustness of the FL approaches to extreme feature heterogeneity in the context of the GEMINI Delirium task. In addition, Appendix \ref{architecture_ablation} reports a final architecture study of the FENDA-FL method highlighting its structural advantage over PerFCL and other approaches.

Throughout the discussion to follow, it is important to note that certain FL approaches require specific model modifications and choices with respect to their setup. For those interested, these changes are thoroughly detailed in Appendix \ref{model_choices} for each approach and dataset. As shown in Table \ref{experimental_model_parameters}, care has been taken to keep the number of trainable parameters approximately constant across different methods.

\subsection{Main Results} \label{main_results}

The main results of this work are exhibited in this section. All experiments use the checkpointing and evaluation strategies detailed in Section \ref{checkpoint_and_eval}. Table \ref{fl_method_rank_table} provides a succinct representation of the overall results. Each method is ranked based on the value achieved for each task and metric considered in the experiments. For approaches with both global and local checkpoints, the metric value of the best performing checkpoint strategy is used to establish rank. Generally, it is evident that personalized approaches provide measurable benefits for the selected tasks over all baselines and non-personalized approaches. FENDA-FL often yields high-performing models and has the highest average rank across all measured metrics, followed closely by Ditto and FedPer. It is important to note that Ditto requires significantly more computational resources, compared with other methods, due to the need for two full-model backpropagation passes at each iteration. Client-side optimizers and tuned hyper-parameters for these experiments are outlined in Appendices \ref{general_experiment_setup}-\ref{gemini_data_hps}. Detailed results for each task are analyzed in the sections below.

\begin{table*}[ht!]
    \centering
    \caption{FL method rankings across all tasks and metrics considered. Non-personalized and personalized FL methods are separated by the horizontal line.}
    \resizebox{0.99\linewidth}{!}{\begin{tabular}{ccccccccc}
    \toprule
     & Fed-Heart & Fed-IXI & Fed-ISIC & \multicolumn{2}{c}{Mortality} & \multicolumn{2}{c}{Delirium} & Avg. Rank\\
    \cmidrule(lr){2-2} \cmidrule(lr){3-3} \cmidrule(lr){4-4} \cmidrule(lr){5-6} \cmidrule(lr){7-8} \cmidrule(lr){9-9}
    Metric & Acc. & Dice & Bal. Acc. & Acc. & AUC/ROC & Acc. & AUC/ROC & \\
    \midrule
    FedAvg & $8$ & $7$ & $3$ & $6$ & $5$ & $6$ & $6$ & $5.86$ \\
    FedAdam & $7$ & $10$ & $5$ & $10$ & $9$ & $9$ & $9$ & $8.43$ \\
    FedProx & $9$ & $8$ & $4$ & $8$ & $7$ & $1$ & $2$ & $5.57$ \\
    SCAFFOLD & $10$ & $9$ & $2$ & $9$ & $2$ & $10$ & $10$ & $7.43$  \\
    MOON & $6$ & $4$ & $7$ & $7$ & 3 & $4$ & $5$ & $5.14$ \\
    \midrule
    FedPer & $2$ & $4$ & $6$ & $2$ & $6$ & $2$ & $3$ & $3.57$ \\
    Ditto & $4$ & $3$ & $1$ & $1$ & $1$ & $7$ & $7$ & $3.43$ \\
    APFL & $5$ & $1$ & $8$ & $5$ & $8$ & $5$ & $4$ & $5.14$ \\
    PerFCL & $3$ & $6$ & $10$ & $4$ & $10$ & $8$ & $8$ & $7.00$ \\
    FENDA-FL & $1$ & $2$ & $9$ & $3$ & $4$ & $3$ & $1$ & $\mathbf{3.29}$ \\
    \bottomrule
    \end{tabular}}
    \label{fl_method_rank_table}
\end{table*}

\subsubsection{FLamby results}

For the FLamby experiments, $15$ server rounds are performed for Fed-Heart-Disease and Fed-ISIC2019, while $10$ are performed for Fed-IXI. Each client completes $100$ local training steps within each round. Training settings for centralized and local training for each dataset are discussed in Appendix \ref{flamby_data_hps}. The loss functions, batch sizes, models, and metrics are all chosen to match the FLamby benchmark settings, unless otherwise specified. The full set of results are displayed in Table \ref{flamby_results_table}.

For Fed-Heart-Disease, FENDA-FL is the best performing approach, followed closely by FedPer. In fact, all personalized FL methods outperform centralized training by a large margin. Furthermore, the personalized methods are the only FL mechanisms to significantly beat siloed training, where each client trains and evaluates on its own data. The non-personalized FL methods largely under-perform. Only the locally checkpointed MOON model produces a mean accuracy above the siloed metric. Finally, the results also suggest that local checkpointing has some advantages for this task, except for SCAFFOLD.

\begin{table*}[ht!]
\centering
    \caption{Performance metrics for the FLamby tasks. Values in parentheses are $95$\% confidence-interval (CI) radii. Global and Local refer to the checkpointing strategies and Silo metrics are measured as described in Section \ref{checkpoint_and_eval}. Bold denotes the best performing FL strategy, while underlined values are second best. Asterisks imply a value is better than siloed training considering CIs. Note that the number of clients varies for each task. For example, Fed-Heart-Disease has only four.}
    \resizebox{\linewidth}{!}{
    \begin{tabular}{lllllll}
        & \multicolumn{2}{c}{Fed-Heart-Disease: Acc.} & \multicolumn{2}{c}{Fed-IXI: Dice} & \multicolumn{2}{c}{Fed-ISIC2019: Balanced Acc.} \\
        \cmidrule(lr){2-3} \cmidrule(lr){4-5} \cmidrule(lr){6-7}
        & \multicolumn{1}{c}{Global} & \multicolumn{1}{c}{Local} & \multicolumn{1}{c}{Global} & \multicolumn{1}{c}{Local} & \multicolumn{1}{c}{Global} & \multicolumn{1}{c}{Local} \\
        \midrule
        FedAvg & $0.724$ ($0.000$) & $0.724$ ($0.000$) & $0.9845^*$ ($0.0001$) & $0.9846^*$ ($0.0001$) & $0.657$ ($0.004$) & $0.644$ ($0.004$) \\
        FedAdam & $0.719$ ($0.000$) & $0.742$ ($0.000$) & $0.9811^*$ ($0.0002$) & $0.9811^*$ ($0.0002$) & $0.647$ ($0.013$) & $0.637$ ($0.012$) \\
        FedProx & $0.716$ ($0.000$) & $0.721$ ($0.000$) & $0.9842^*$ ($0.0001$) & $0.9844^*$ ($0.0001$) & $0.633$ ($0.022$) & $0.652$ ($0.022$) \\
        SCAFFOLD & $0.711$ ($0.000$) & $0.682$ ($0.000$) & $0.9827^*$ ($0.0000$) & $0.9827^*$ ($0.0000$) & $\underline{0.672}^*$ ($0.013$) & $0.669$ ($0.010$) \\
        MOON & $0.727$ ($0.010$) & $0.752$ ($0.036$) & $0.9852^*$ ($0.0002$) & $0.9852^*$ ($0.0002$) & $0.623$ ($0.021$) & $0.605$ ($0.015$) \\
        FedPer & \multicolumn{1}{c}{--} & $\underline{0.814}^*$ ($0.002$) & \multicolumn{1}{c}{--} & $0.9852^*$ ($0.0001$) & \multicolumn{1}{c}{--} & $0.634$ ($0.014$) \\
        {Ditto} & \multicolumn{1}{c}{--} & $0.802^*$ ($0.005$) & \multicolumn{1}{c}{--} & $0.9854^*$ ($0.0002$) & \multicolumn{1}{c}{--} & $\mathbf{0.703}^*$ ($0.009$) \\
        APFL & \multicolumn{1}{c}{--} & $0.801^*$ ($0.006$) & \multicolumn{1}{c}{--} & $\mathbf{0.9864}^*$ ($0.0002$) & \multicolumn{1}{c}{--} & $0.608$ ($0.011$) \\
        PerFCL & \multicolumn{1}{c}{--} & $0.805^*$ ($0.018$) & \multicolumn{1}{c}{--} & $0.9847^*$ ($0.0001$) & \multicolumn{1}{c}{--} & $0.584$ ($0.019$) \\
        FENDA-FL & \multicolumn{1}{c}{--} & $\mathbf{0.815}^*$ ($0.000$) & \multicolumn{1}{c}{--} & $\underline{0.9856}^*$ ($0.0001$) & \multicolumn{1}{c}{--} & $0.607$ ($0.011$) \\
        \midrule
        Silo & \multicolumn{1}{c}{--} & $0.748$ ($0.000$) & \multicolumn{1}{c}{--} & $0.9737$ ($0.0040$) & \multicolumn{1}{c}{--} & $0.634$ ($0.025$) \\
        Central & \multicolumn{1}{c}{--} & $0.732$ ($0.000$) & \multicolumn{1}{c}{--} & $0.9832$ ($0.0002$) & \multicolumn{1}{c}{--} & $0.683$ ($0.020$) \\
        Client 0 & \multicolumn{1}{c}{--} & $0.735$ ($0.000$) & \multicolumn{1}{c}{--} & $0.9648$ ($0.0123$) & \multicolumn{1}{c}{--} & $0.549$ ($0.014$) \\
        Client 1 & \multicolumn{1}{c}{--} & $0.649$ ($0.000$) & \multicolumn{1}{c}{--} & $0.9722$ ($0.0004$) & \multicolumn{1}{c}{--}  & $0.387$ ($0.029$) \\
        Client 2 & \multicolumn{1}{c}{--} & $0.589$ ($0.000$) & \multicolumn{1}{c}{--} & $0.9550$ ($0.0015$) & \multicolumn{1}{c}{--} & $0.554$ ($0.010$) \\
        Client 3 & \multicolumn{1}{c}{--} & $0.629$ ($0.000$) & \multicolumn{1}{c}{--} & \multicolumn{1}{c}{--} & \multicolumn{1}{c}{--} & $0.444$ ($0.016$) \\
        Client 4 & \multicolumn{1}{c}{--} & \multicolumn{1}{c}{--} & \multicolumn{1}{c}{--} & \multicolumn{1}{c}{--} & \multicolumn{1}{c}{--} & $0.326$ ($0.018$) \\
        Client 5 & \multicolumn{1}{c}{--} & \multicolumn{1}{c}{--} & \multicolumn{1}{c}{--} & \multicolumn{1}{c}{--} & \multicolumn{1}{c}{--} & $0.285$ ($0.039$) \\
        \bottomrule
    \end{tabular}}
    \label{flamby_results_table}
\end{table*}

As seen in Table \ref{flamby_results_table}, each method produces a model that performs the Fed-IXI task well. However, there is a clear advantage to central training compared with individual client or siloed training. An even larger improvement is realized with FL. Among non-personalized FL methods, models trained with FedAvg, FedProx, and MOON surpass central training. All personalized FL approaches also improve upon the centrally trained model and produce four of the top five best metrics. APFL and FENDA-FL yield the best models, with APFL beating FENDA-FL by a small margin. Importantly, additional experiments, discussed in Appendix \ref{architecture_ablation}, show that this gap can be closed with thoughtful architecture design for FENDA-FL. Finally, the performance variance for federally trained models is equivalent to the centrally trained version. As with Fed-Heart-Disease, using local checkpointing has small benefits for most methods.

Fed-ISIC2019 is a particularly challenging task. The results in Table \ref{flamby_results_table} show that all FL strategies perform better than local models, with Ditto providing impressive results. It is the only method to surpass centralized training, though some other methods approach the performance. Most of the non-personalized FL methods equal or surpass siloed evaluation, though only SCAFFOLD's model is statistically significant. Alternatively, for this task, all personalized FL approaches, with the stark exception of Ditto, under-perform. While the results are better than any of the local models, non-personalized methods, with the exception of MOON, consistently out-perform these techniques. Moreover, they fail to eclipse siloed accuracy. An investigation into where these approaches fall short for this dataset appears in Appendix \ref{fed_isic_analysis}. It is worth noting that, for this task, Ditto required significantly more compute resources than any other method, but also achieves impressive accuracy.
 
\paragraph{FLamby Benchmark Improvements}

Beyond the addition of personalized FL methods to the FLamby benchmark, the overall results achieved for Fed-IXI and FedISIC2019 stand in contrast to those reported in the original work. Therein, all FL strategies severely under-performed even local training for Fed-IXI. Similarly, none of the FL methods showed strong performance for Fed-ISIC2019 in the earlier study. This led the authors to conclude that federated training was not a useful approach for these tasks. The results presented here compellingly support the opposite conclusion.

In the original study, a standard SGD optimizer is used for client-side training, whereas AdamW is used here. Furthermore, the models for Fed-IXI and Fed-ISIC2019 employ Batch Normalization layers. When using FedAdam, it is critical that stats tracking be turned off for these layers to avoid issues during the aggregation process. This phenomenon is discussed in Appendix \ref{momentum_based_optimizers_and_bn}. We speculate that a related issue may have impeded the convergence of the FedOpt family in their experiments.

\subsubsection{GEMINI Results}

For both GEMINI tasks, $50$ server rounds are performed. The clients in Mortality and Delirium prediction complete $2$ and $4$ local epochs, respectively, at each server round. GEMINI training setups, hyper-parameters, and other details are discussed in Appendix \ref{gemini_data_hps}. Models for each task are assessed based on accuracy and AUC/ROC.

\begin{table*}[ht!]
    \centering
    \caption{Performance metrics for GEMINI Mortality. Values in parentheses are $95$\% confidence-interval (CI) radii. Global and Local refer to the checkpointing strategies and Silo metrics are measured as described in Section \ref{checkpoint_and_eval}. Bold denotes the best performing federated strategy, while underlined values are second best. Asterisks imply a value is better than siloed training considering CIs.}
    \resizebox{0.75\linewidth}{!}{
    \begin{tabular}{lllll}
        &\multicolumn{2}{c}{Mean Accuracy} & \multicolumn{2}{c}{Mean AUC/ROC}\\
        \cmidrule(lr){2-3} \cmidrule(lr){4-5}
         & \multicolumn{1}{c}{Global} & \multicolumn{1}{c}{Local} & \multicolumn{1}{c}{Global} & \multicolumn{1}{c}{Local} \\
        \midrule
        FedAvg & $0.8976$ ($0.0002$) & $0.8971$ ($0.0006$) & $0.8180^*$ ($0.0005$) & $0.8177^*$ ($0.0005$)\\
        FedAdam & $0.8954$ ($0.0000$) & $0.8954$ ($0.0000$) & $0.8128$ ($0.0005$) & $0.8123$ ($0.0008$) \\
        FedProx & $0.8957$ ($0.0004$) & $0.8956$ ($0.0003$) & $0.8174^*$ ($0.0004$) & $0.8172^*$ ($0.0006$) \\
        SCAFFOLD & $0.8955$ ($0.0003$) & $0.8954$ ($0.0000$) & $\underline{0.8186}^*$ ($0.0002$) & $0.8184^*$ ($0.0002$) \\
        MOON & $0.8968$ ($0.0008$) & $0.8970$ ($0.0002$) & $0.8184^*$ ($0.0003$) & $0.8181^*$ ($0.0008$) \\
        FedPer & \multicolumn{1}{c}{--} & $\underline{0.8985}$ ($0.0005$) & \multicolumn{1}{c}{--} & $0.8179^*$ ($0.0006$) \\
        Ditto & \multicolumn{1}{c}{--} & $\mathbf{0.8994}^*$ ($0.0003$) & \multicolumn{1}{c}{--} &$\mathbf{0.8201}^*$ ($0.0011$) \\
        APFL & \multicolumn{1}{c}{--} & $0.8978$ ($0.0004$) & \multicolumn{1}{c}{--} & $0.8138$ 
        ($0.0017$) \\
        PerFCL & \multicolumn{1}{c}{--} & $0.8979$ ($0.0008$) & \multicolumn{1}{c}{--} & $0.8117$ ($0.0012$) \\
        FENDA-FL & \multicolumn{1}{c}{--} & $0.8980$ ($0.0004$) & \multicolumn{1}{c}{--} & $0.8181^*$ ($0.0001$) \\
        \midrule
        Silo & \multicolumn{1}{c}{--} & $0.8978$ ($0.0009$) & \multicolumn{1}{c}{--} & $0.8140$ ($0.0012$)   \\
        Central & \multicolumn{1}{c}{--} & $0.8981$ ($0.0005$) & \multicolumn{1}{c}{--} & $0.8193$ ($0.0005$) \\
        Client 0 & \multicolumn{1}{c}{--} & $0.8966$ ($0.0014$) & \multicolumn{1}{c}{--} & $0.8052$ ($0.0030$) \\
        Client 1 & \multicolumn{1}{c}{--} & $0.8930$ ($0.0025$) & \multicolumn{1}{c}{--} & $0.7657$ ($0.0064$) \\
        Client 2 & \multicolumn{1}{c}{--} & $0.8943$ ($0.0060$) & \multicolumn{1}{c}{--} & $0.7591$ ($0.0116$) \\
        Client 3 & \multicolumn{1}{c}{--} & $0.8954$ ($0.0000$) & \multicolumn{1}{c}{--} & $0.7994$ ($0.0034$) \\
        Client 4 & \multicolumn{1}{c}{--} & $0.8955$ $(0.0005)$ & \multicolumn{1}{c}{--} & $0.7974$ ($0.0037$) \\
        Client 5 & \multicolumn{1}{c}{--} & $0.8973$ $(0.0014)$ & \multicolumn{1}{c}{--} & $0.7878$ ($0.0043$) \\
        Client 6 & \multicolumn{1}{c}{--} & $0.8954$ $(0.0000)$ & \multicolumn{1}{c}{--} & $0.8019$ ($0.0035$) \\
        \bottomrule
        \end{tabular}}
        \label{mortality_results_table}
\end{table*}

\begin{table*}[ht!]
    \centering
    \caption{Performance metrics for GEMINI Delirium. Values in parentheses are $95$\% confidence-interval (CI) radii. Global and Local refer to the checkpointing strategies and Silo metrics are measured as described in Section \ref{checkpoint_and_eval}. Bold denotes the best performing federated strategy, while underlined values are second best. Asterisks imply a value is better than siloed training considering CIs.}
    \resizebox{0.75\linewidth}{!}{
    \begin{tabular}{lllll}
        & \multicolumn{2}{c}{Mean Accuracy} & \multicolumn{2}{c}{Mean AUC/ROC} \\
        \cmidrule(lr){2-3} \cmidrule(lr){4-5}
        & \multicolumn{1}{c}{Global} & \multicolumn{1}{c}{Local} & \multicolumn{1}{c}{Global} & \multicolumn{1}{c}{Local} \\
        \midrule
        FedAvg & $0.7987$ ($0.0127$) & $0.8025$ ($0.0130$) & $0.8302^*$ ($0.0146$) & $0.8274^*$ ($0.0078$) \\
        FedAdam & $0.7689$ ($0.0036$) & $0.7688$ ($0.0127$) & $0.7897$ ($0.0078$) & $0.7881$ ($0.0055$) \\
        FedProx & $\mathbf{0.8095}^*$ ($0.0036$) & $0.8056$ ($0.0037$) & $0.8488^*$ ($0.0043$) & $\underline{0.8504}^*$ ($0.0042$) \\
        SCAFFOLD &  $0.7480$ ($0.0000$) & $0.7480$ ($0.0000$) & $0.5491$ ($0.0469$) & $0.5491$ ($0.0469$) \\
        MOON &  $0.8056$ ($0.0100$) & $0.8006$ ($0.0140$) & $ 0.8375^*$ ($0.0051$) & $0.8365^*$ ($0.0027$) \\
        FedPer & \multicolumn{1}{c}{--}  & $\underline{0.8082}^*$ ($0.0024$) & \multicolumn{1}{c}{--} & $0.8462^*$ ($0.0035$) \\
        Ditto & \multicolumn{1}{c}{--} & $0.7981$ ($0.0034$) & \multicolumn{1}{c}{--} & $0.8271^*$ ($0.0017$) \\
        APFL & \multicolumn{1}{c}{--}  & $0.8031$ ($0.0052$) & \multicolumn{1}{c}{--}  & $0.8430^*$ ($0.0062$) \\
        PerFCL & \multicolumn{1}{c}{--}  & $0.7893$ ($0.0077$) & \multicolumn{1}{c}{--}  & $0.8111$ ($0.0089$) \\
        FENDA-FL & \multicolumn{1}{c}{--}  & $0.8064$ ($0.0036$) & \multicolumn{1}{c}{--}  & $\mathbf{0.8518}^*$ ($0.0017$) \\
        \midrule
        Silo & \multicolumn{1}{c}{--}  & $0.7936$ ($0.0102$) &\multicolumn{1}{c}{--}  & $0.8037$ ($0.0072$) \\
        Central & \multicolumn{1}{c}{--}  & $0.8114$ ($0.0068$) & \multicolumn{1}{c}{--} & $0.8458$ ($0.0026$) \\
        Client 0 & \multicolumn{1}{c}{--}  & $0.7653$ ($0.0070$) & \multicolumn{1}{c}{--} & $0.7977$ ($0.0128$) \\
        Client 1 & \multicolumn{1}{c}{--} & $0.7781$ ($0.0063$) & \multicolumn{1}{c}{--} & $0.7795$ ($0.0161$)\\
        Client 2 & \multicolumn{1}{c}{--} & $0.7805$ ($0.0036$) & \multicolumn{1}{c}{--} & $0.8185^*$ ($0.0021$)\\
        Client 3 & \multicolumn{1}{c}{--} & $0.7634$ ($0.0062$) & \multicolumn{1}{c}{--} & $0.7539$ ($0.0055$)\\
        Client 4 & \multicolumn{1}{c}{--} & $0.7924$ ($0.0061$) & \multicolumn{1}{c}{--} & $0.8179$ ($0.0076$)\\
        Client 5 & \multicolumn{1}{c}{--} & $0.8028$ ($0.0040$) & \multicolumn{1}{c}{--} & $0.8429^*$ ($0.0018$)\\
        \bottomrule
    \end{tabular}}
    \label{delirium_results_table}
\end{table*}

The results for the GEMINI Mortality task are shown in Table \ref{mortality_results_table}. FL methods improve upon locally trained models, reducing the performance gap to central training. This is especially true when considering AUC/ROC. In terms of accuracy, FedAvg is the only standard method that rivals the personalized FL algorithms and approaches siloed evaluation. On the other hand, most methods surpass the siloed AUC/ROC evaluation metric, nearing centrally trained model performance. Specifically, FedProx, SCAFFOLD, and MOON beat siloed training based on AUC/ROC, but did not in terms of accuracy. Ditto is the best approach, outperforming siloed and central accuracy and AUC/ROC. Considering both metrics, FedPer and FENDA-FL also provide strong results, each surpassing the siloed evaluation metrics. Both methods are competitive with non-personalized approaches in terms of AUC/ROC and outperform them based on accuracy.

In the GEMINI Delirium results, reported in Table \ref{delirium_results_table}, non-personalized approaches such as FedProx, MOON, and FedAvg outperform siloed evaluation in both metrics, while SCAFFOLD and FedAdam under-perform, despite an extensive hyper-parameter sweep. FedPer, APFL, and FENDA-FL also outperform the siloed evaluation metrics. Notably, both PerFCL and Ditto, the best performing method for the Mortality task, fail to outperform several non-personalized FL methods in terms of accuracy and AUC/ROC. Across all approaches, based on accuracy, only FedProx’s globally checkpointed model and FedPer surpass the performance of FENDA-FL, while it is the best method in terms of AUC/ROC. 

Interestingly, locally trained models from Client 2 and Client 5, surpass siloed evaluation in AUC/ROC. These clients have the largest number of samples, see Table \ref{fl_dataset_distributions}. Client 5's model appears to be particularly generalizable. For example, it outperforms the local models of Clients 1, 2, 3, and 4 evaluated on their own test data, sometimes by a wide margin. As such, the improvements achieved by FENDA-FL are notable.

\subsection{Federated Checkpointing Ablation Study}
\label{checkpoint_ablation}

To quantify the benefits of the federated checkpointing strategies proposed in Section \ref{checkpoint_and_eval}, the resulting models are compared against the widely used fixed-rounds strategy, where only the model from the final FL round is saved. For non-personalized approaches, this is the final aggregated model on the server. For personalized approaches, the final-round client-side model is saved after any server-side aggregation. These checkpoints are referred to as ``Latest'' in the results shown in Table \ref{checkpoint_ablation_results}.

For Fed-Heart-Disease, there is a clear advantage to federated checkpointing, especially the local version, for most strategies. Notably, APFL and FENDA-FL benefit significantly. For Fed-IXI, two experiments are conducted. In the first, optimal hyper-parameters for each strategy are applied during training. The benefits of local checkpointing are less pronounced in this setting, but some of the approaches exhibit small improvements. APFL, PerFCL, and FENDA see larger improvements. 

To consider settings when the optimal number of rounds is not well calibrated, training with sub-optimal hyper-parameters, detailed in Appendix \ref{sub_optimal_fed_ixi}, is also performed. In this setting, a significant advantage is seen when using federated checkpointing, with the exception of PerFCL and Ditto. Local federated checkpointing also outperforms the server-side analogue. Federated checkpointing also results in a measurable reduction in variance for all strategies, again excepting PerFCL and Ditto. The performance drop and variance increase for PerFCL and Ditto can be traced to a single federated run failing to converge during training, yielding a single instance of very poor performance. In both cases, if the divergent run is excluded, federated checkpointing again produces similar or superior results.

\begin{table*}[ht!]
    \centering
    \caption{Checkpoint ablation results comparing federated checkpointing strategies with saving models after a fixed number of FL rounds. Values in parentheses are $95$\% confidence-interval radii.}
    \resizebox{\linewidth}{!}{\begin{tabular}{lllllll}
        & \multicolumn{3}{c}{Fed-Heart-Disease: Mean Accuracy} & \multicolumn{3}{c}{Fed-IXI Optimal Parameters: Mean Dice} \\
        \midrule
        & \multicolumn{1}{c}{Latest} & \multicolumn{1}{c}{Global} & \multicolumn{1}{c}{Local} & \multicolumn{1}{c}{Latest} & \multicolumn{1}{c}{Global} & \multicolumn{1}{c}{Local} \\
        \cmidrule(lr){2-2} \cmidrule(lr){3-3} \cmidrule(lr){4-4} \cmidrule(lr){5-5} \cmidrule(lr){6-6} \cmidrule(lr){7-7}
        FedAvg & $0.706$ ($0.000$) & $\mathbf{0.724}$ ($0.000$) & $\mathbf{0.724}$ ($0.000$) & $0.9845$ ($0.0002$) & $0.9845$ ($0.0001$) & $\mathbf{0.9846}$ ($0.0001$) \\
        FedAdam & $0.711$ ($0.000$) & $0.719$ ($0.000$) & $\mathbf{0.742}$ ($0.000$) & $\mathbf{0.9815}$ ($0.0001$) & $0.9811$ ($0.0002$) & $0.9811$ ($0.0002$) \\
        FedProx & $0.703$ ($0.000$) & $0.716$ ($0.000$) & $\mathbf{0.721}$ ($0.000$) & $0.9841$ ($0.0001$) & $0.9842$ ($0.0001$) & $\mathbf{0.9844}$ ($0.0001$) \\
        SCAFFOLD & $\mathbf{0.717}$ ($0.000$) & $0.711$ ($0.000$) & $0.682$ ($0.000$) & $\mathbf{0.9828}$ ($0.0000$) & $0.9827$ ($0.0000$) & $0.9827$ ($0.0000$) \\
        MOON & $0.725$ ($0.009$) & $0.727$ ($0.010$) & $\mathbf{0.752}$ ($0.036$) & $0.9851$ ($0.0002$) & $\mathbf{0.9852}$ ($0.0002$) & $\mathbf{0.9852}$ ($0.0002$) \\
        FedPer & $\mathbf{0.814}$ ($0.003$) & \multicolumn{1}{c}{--} & $\mathbf{0.814}$ ($0.002$) & $\mathbf{0.9852}$ ($0.0003$) & \multicolumn{1}{c}{--} & $0.9847$ ($0.0001$) \\
        Ditto & $0.797$ ($0.007$) & \multicolumn{1}{c}{--} & $\mathbf{0.802}$ ($0.005$) & $\mathbf{0.9854}$ ($0.0003$) & \multicolumn{1}{c}{--} & $\mathbf{0.9854}$ ($0.0002$) \\
        APFL & $0.777$ ($0.013$) & \multicolumn{1}{c}{--} & $\mathbf{0.801}$ ($0.006$) & $0.9858$ ($0.0000$) & \multicolumn{1}{c}{--} & $\mathbf{0.9864}$ ($0.0002$) \\
        PerFCL & $\mathbf{0.806}$ ($0.010$) & \multicolumn{1}{c}{--} & $0.805$ ($0.018$) & $0.9843$ ($0.0004$) & \multicolumn{1}{c}{--} & $\mathbf{0.9847}$ ($0.0001$) \\
        FENDA-FL & $0.802$ ($0.010$) & \multicolumn{1}{c}{--} & $\mathbf{0.815}$ ($0.000$) & $0.9848$ ($0.0026$) & \multicolumn{1}{c}{--} & $\mathbf{0.9856}$ ($0.0001$) \\
        & & & & & & \\
        & \multicolumn{3}{c}{Fed-IXI Sub-Optimal Parameters: Mean Dice} & &\\
        \cmidrule(lr){1-4}
        & \multicolumn{1}{c}{Latest} & \multicolumn{1}{c}{Global} & \multicolumn{1}{c}{Local}  &  & &\\
         \cmidrule(lr){2-2} \cmidrule(lr){3-3} \cmidrule(lr){4-4}
         FedAvg & $0.9766$ ($0.0006$) & $\mathbf{0.9788}$ ($0.0006$) & $\mathbf{0.9788}$ ($0.0006$) &  & &\\
        FedAdam & $0.9202$ ($0.0091$) & $0.9483$ ($0.0056$)  & $\mathbf{0.9488}$ ($0.0054$)  &  & &\\
        FedProx & $0.9295$ ($0.0356$) & $0.9548$ ($0.0131$) & $\mathbf{0.9558}$ ($0.0136$) & & &\\
        SCAFFOLD & $\mathbf{0.4360}$ ($0.0000$) & $\mathbf{0.4360}$ ($0.0000$) & $\mathbf{0.4360}$ ($0.0000$) &  & &\\
        MOON & $0.9177$ ($0.1149$) & $0.9769$ ($0.0012$) & $\mathbf{0.9771}$ ($0.0013$) &  & &\\
        FedPer & $0.9754$ ($0.0019$) & \multicolumn{1}{c}{--} & $\mathbf{0.9779}$ ($0.0015$) & & & \\
        Ditto & $\mathbf{0.9789}$ ($0.0014$) & \multicolumn{1}{c}{--} & $0.9757$ ($0.0083$) & & & \\
        APFL & $0.9780$ ($0.0007$) & \multicolumn{1}{c}{--} & $\mathbf{0.9793}$ ($0.0005$) & & & \\
        PerFCL & $\mathbf{0.9853}$ ($0.0005$) & \multicolumn{1}{c}{--} & $0.9711$ ($0.0282$) & & & \\
        FENDA-FL & $0.9705$ ($0.0075$) & \multicolumn{1}{c}{--} & $\mathbf{0.9790}$ ($0.0004$) & & & \\
    \end{tabular}}
    \label{checkpoint_ablation_results}
\end{table*}

\subsection{Extreme Heterogeneity Study}
\label{delirium_het_study}

\begin{figure*}[ht!]
    \centering
    \includegraphics[scale=0.14]{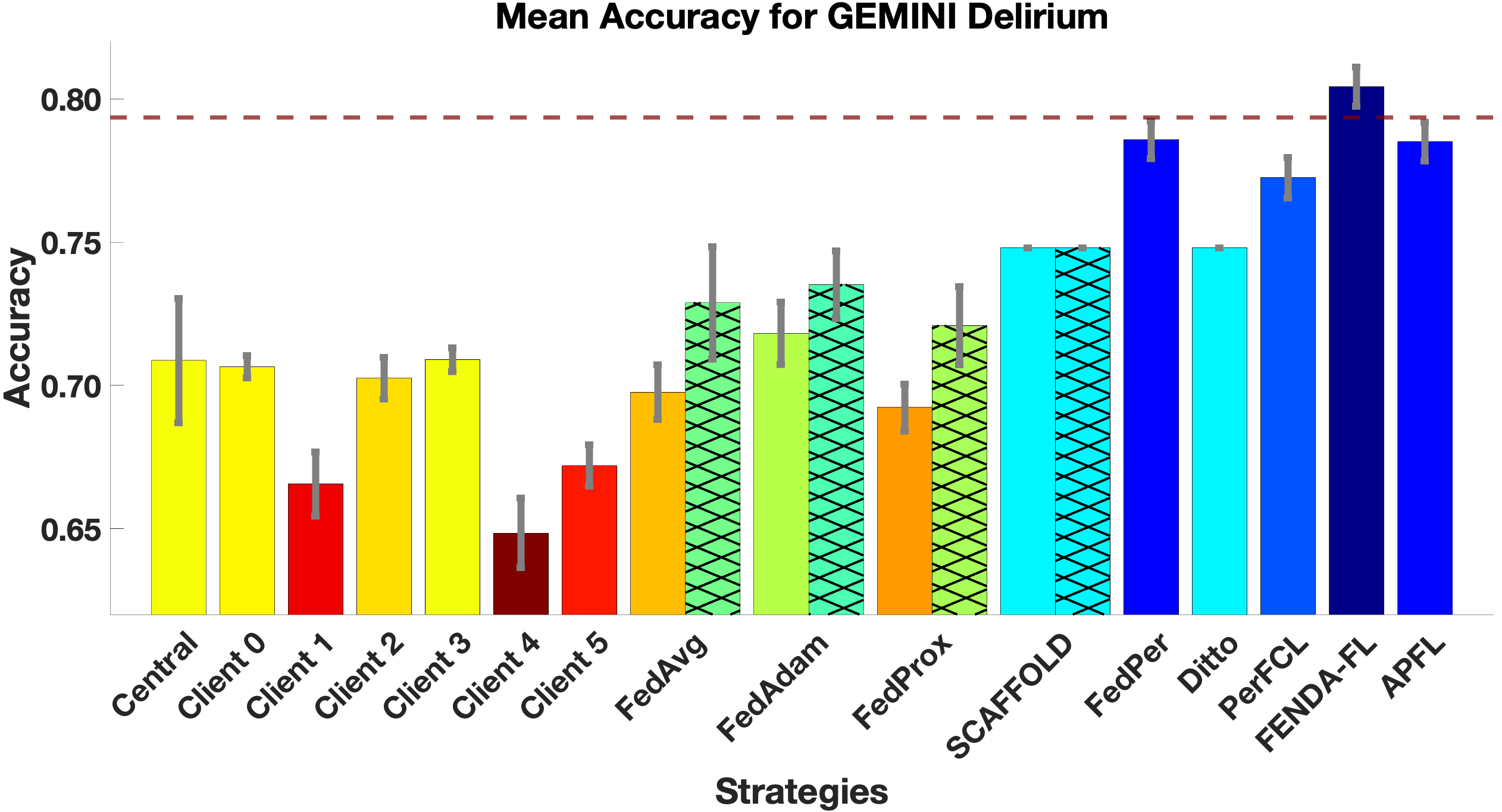}
    \caption{Mean accuracy across clients for the GEMINI Delirium task when each client performs independent feature processing. The dashed horizontal line represents siloed performance, see Section \ref{checkpoint_and_eval}. Solid and hatched bars indicate global and local checkpointing, respectively.}
    \label{delirium_extreme_heterogeneity}
\end{figure*}

As a test of the robustness of the FL methods, an experiment inducing extreme feature-space heterogeneity is designed. 
For the GEMINI Delirium task, described in Section \ref{delirium_dataset}, each client locally processes its data with dimensionality reduction through PCA and feature selection to produce input vectors of dimension $300$. As clients independently perform this process, the resulting feature spaces are no longer well-aligned. In this setup, locally trained models are unlikely to generalize well to other domains. Moreover, FL techniques that produce a single shared model are expected to encounter convergence issues. This is confirmed in the results shown in Figure \ref{delirium_extreme_heterogeneity}. The results highlight the robustness and effectiveness of personalized methods. In particular, FENDA-FL demonstrates notable performance in this setting as the only method to recover and surpass the siloed performance target. Note also that local checkpointing is important for this task.

\section{Discussion}

In this work, we make substantive contributions to FL research with a specific focus on challenging clinical applications. The experiments represent an extension of the FLamby benchmark along two dimensions for the selected tasks. First, the results show that FL methods provide convincing utility for all tasks, including Fed-IXI and Fed-ISIC2019, improving on the original benchmark experiments. Furthermore, the reported results provide strong personalized FL baselines not present in the original study. The GEMINI results offer another conclusive validation of FL models, especially personalized approaches, in true clinical settings and expand the datasets studied. In both cases, personalized FL models offer marked improvements in task performance over non-personalized techniques, as seen in Table \ref{fl_method_rank_table}. Across various tasks and metrics, personalized techniques are even capable of significantly outperforming centralized training. Overall, these results demonstrate the significant potential of personalized FL methods to improve clinical modeling by balancing global learning with site-specific predictions.

In addition, we outline a comprehensive and effective checkpointing and evaluation framework, building on work in the original FLamby benchmark. This framework better aligns federated training with standard ML practices and produces reliable measures of FL model performance. Experiments demonstrate the utility of the checkpointing approaches. Finally, an important ablation of PerFCL, FENDA-FL, is investigated and experiments demonstrate its effectiveness for clinical tasks and robustness to heterogeneous distributions. Notably, it outperforms PerFCL and other personalized approaches in many cases. Auxiliary studies, reported in the appendices, demonstrate the benefit of FENDA-FL's architectural flexibility in optimizing performance. Altogether, this work provides consequential insights into the utility of FL methods, especially personalized techniques, in the context of real-world clinical tasks.

\paragraph{Limitations}

While the benchmarking and other experimental results presented are extensive, other personalized and non-personalized methods exist. We aim to continue to expand this work and the associated library to include emerging approaches. Additionally, the mechanisms of failure and strategies for improving the performance of most of the personalized methods on the Fed-ISIC2019 task require further investigation. In future work, we plan to consider the application of warm-start strategies and auxiliary losses as one avenue for improvement. Finally, the interplay between model generalizability and personalization remains under-explored. To this end, this work will be extended to consider domain generalization techniques such as Generalization Adjustment \citep{zhang2023federated}.

\section{Conclusion}

In this paper, we have extensively studied personalized and non-personalized FL methods in the context of challenging real-world clinical tasks. Overall, FL methods, especially personalized approaches, prove to be extremely effective at improving performance across each of the studied datasets. As part of this benchmarking effort, we have introduced a comprehensive framework for federated evaluation, aligning the process more closely with standard ML workflows. Finally, we studied an important ablation of PerFCL, FENDA-FL. The approach yields the best mean rank and produces good performance across most tasks.




\bibliography{custom}

\appendix

\section{Choice of Personalized FL Methods} \label{pfl_method_choice}

A careful selection of state-of-the-art personalized FL approaches in the context of clinical tasks is studied. PerFCL \citep{Zhang1} is selected as a personalized extension of MOON \citep{Li2}, a recent non-personalized FL method. Ditto \citep{Ditto1} is included as it showed solid benchmark performance in \citep{Matsuda1}. APFL \citep{Deng1} is a common benchmark method, especially in non-IID settings. Moreover, it shares architectural similarities with PerFCL and FENDA-FL, making it a particularly good baseline. Finally, FedPer is chosen as a sequential personalized model, in contrast to the parallel models of APFL, PerFCL, and FENDA-FL. It is also chosen as representative of FedRep, a similar approach with nearly identical benchmark performance in \citep{Matsuda1} and a strong competitor to FedPop in \citep{Kotelevski1}.

Due to incompatible settings or narrow architectures, several well-known methods are excluded. FedBN \citep{Li1} is restricted to models with batch normalization layers. \citet{Jeong1} consider settings of extreme label and domain heterogeneity not present here. FedME \citep{Matsuda2}, the best performing method in \citep{Matsuda1}, assumes that the server holds a relevant, generalizable dataset to facilitate clustering, which is often infeasible in clinical settings. Finally, HypCluster \citep{Mansour1} partitions clients into clusters and trains models for each cluster. When the number of clients is small, as in this work, such partitioning is impractical and forgoes global information.

In addition, various methods are excluded due to poor benchmark performance. Both FML \citep{Shen2} and LG-FedAvg \citep{Liang1} performed poorly on the \citep{Matsuda1} benchmark. Similarly, pFedMe \citep{Dinh1} failed to outperform FedProx in a number of settings \citep{Matsuda1, Chen1}. Finally, we do not consider FedPop \citep{Kotelevski1} as it is explicitly designed for cross-device settings rather than cross-silo FL. Therein, the experimental results utilize a minimum of $100$ clients with small local datasets.

Other personalized FL approaches exist. For example, FedEM \citep{Marfoq1} shows promise in \citep{Chen1}, though in some settings still struggles to outperform FedProx. While the results presented here are not exhaustive, they provide an important benchmark of personalized and non-personalized FL methods for clinical tasks.

\section{Model Architectures for Each Task}
\label{model_choices}

Table \ref{experimental_model_parameters} provides a comparison of the trainable parameters present in the models used in the FLamby and GEMINI experiments. The ``standard'' column refers to models trained in all settings, excluding APFL, FENDA-FL, and PerFCL. Note that FedPer and MOON use the $151$ parameter models for Fed-Heart-Disease to facilitate feature extraction. The number of trainable parameters across different methods has been kept approximately constant to ensure that different methods do not benefit from out-sized expressiveness.

\begin{table*}[ht!]
    \centering
    \caption{Trainable parameters for the models used in the experiments. Standard refers to models used in all settings outside of APFL, FENDA-FL, and PerFCL.}
    \begin{tabular}{lllll}
        \toprule
        Dataset & Standard & APFL & FENDA-FL & PerFCL \\
        \midrule
        Fed-Heart-Disease & $14$ or $151$ & $152$ & $151$ & $151$ \\
        \midrule
        Fed-IXI & $1,106,520$ & $984,624$ & $984,620$ & $984,620$ \\
        \midrule
        Fed-ISIC2019 & $4,017,796$ & $4,631,088$ & $4,775,016$ & $4,775,016$ \\
        \midrule
        Mortality & $222,784$ & $222,784$ & $222,784$ & $222,784$ \\
        \midrule
        Delirium & $8,988,736$ & $8,988,736$ & $8,717,448$ & $8,717,448$ \\
        \bottomrule
    \end{tabular}
    \label{experimental_model_parameters}
\end{table*}

\subsection{Fed-Heart-Disease}

The baseline model for Fed-Heart-Disease is logistic regression with a feature space of only $13$ dimensions. However, a deeper model is required for several approaches to implement the strategy. Both MOON and FedPer use a single linear layer and ReLU activation for their feature extractors. For PerFCL and FENDA-FL, both feature extractors are also single linear layers with a ReLU activation. For each approach, the classification head is a linear layer followed by a sigmoid. For APFL, a two-layer dense neural net (DNN) is used for both the global and local model with a ReLU in between followed by a sigmoid. 

\begin{figure*}[ht!]
    \centering
    \includegraphics[scale=0.145]{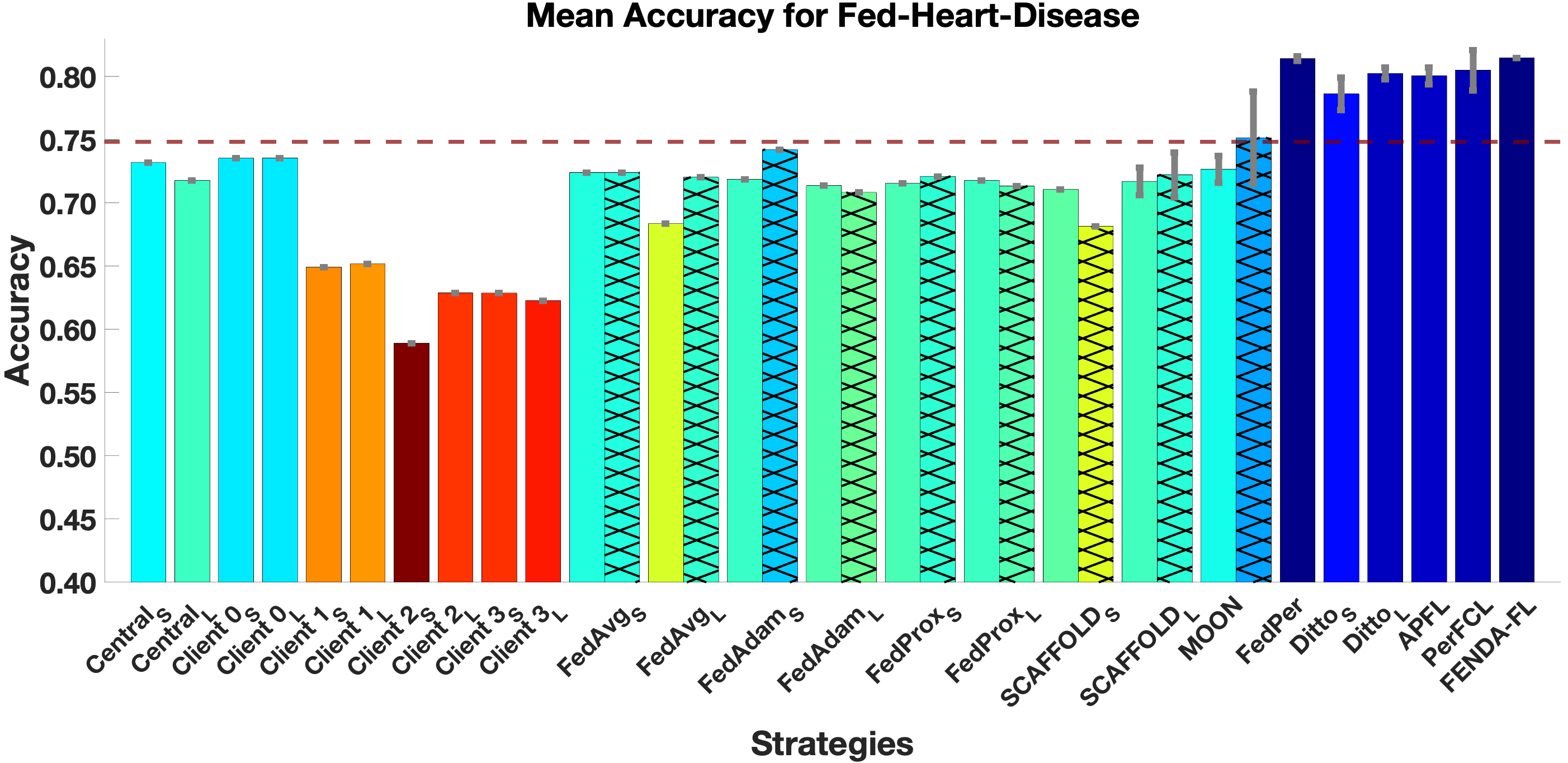}
    \caption{Mean accuracy for the Fed-Heart-Disease task. This figure shows results for small and large models with $14$ and $151$ parameters, where applicable. Strategies sub-scripted with an S use small models and those with an L use the large models. Those without subscripts are large-model results. Solid and hatched bars indicate global and local checkpointing, respectively.}
    \label{fed_heart_disease_all_models}
\end{figure*}

Figure \ref{fed_heart_disease_all_models} displays the performance of both small and large models for various training strategies on Fed-Heart-Disease. The small models are logistic regression with only $14$ trainable parameters. The large versions are two-layer DNNs with $151$ parameters. The models used by FENDA-FL/PerFCL and APFL are DNN-based with $151$ and $152$ parameters, respectively. For strategies other than FENDA-FL, PerFCL, and APFL, the large model variants generally do not improve performance, with the exception of Ditto. As such, performance of the large model is reported for Ditto in the main results. Note that federated checkpointing, as described in Section \ref{checkpoint_and_eval}, is applied in all cases. Therefore, it is unlikely that this phenomenon is due to over-fitting.

\subsection{Fed-IXI}
As a 3D-segmentation task, the metric for Fed-IXI is a Dice score. The model architecture is a 3D U-Net \citep{Cicek1} and the implementation in FLamby is used. The encoder produces voxel-wise features of the 3D volume at different resolutions that are concatenated along the channel dimension with features from the decoder of the same resolution. The per-voxel channel contents of features at the highest resolution are then used to produce a binary classification for each voxel. For a fair comparison to APFL, PerFCL, and FENDA-FL, which require architecture modifications, the number of such channels in the U-Net is increased to $12$ for other methods, compared with the FLamby default of $8$. The models used for APFL, PerFCL, and FENDA-FL each produce $8$ channels. PerFCL and FENDA-FL models use the U-Net without a classifier for both the global and local modules. The resulting activations are concatenated and fed into a classification layer. The FedPer and MOON models are split between the U-Net base and the voxel-wise binary classifier.

\subsection{Fed-ISIC2019}
The model used for Fed-ISIC2019 is EfficientNet-B0 \citep{Tan1}, which has been pretrained on ImageNet, with a linear final layer. The classifier layer for MOON and FedPer is defined as the last linear layer and the remainder of the model is the feature extractor. For APFL, the full model is used for the global and local modules. Finally, the PerFCL and FENDA-FL models remove the final layer and use the remaining architecture as the local and global feature extractors. The classification head is a two-layer DNN with a ReLU between the layers. For a fair comparison, the first $13$ layers of EfficientNet-B0 are frozen for APFL, PerFCL, and FENDA-FL to limit the trainable parameters.

\subsection{Mortality}
The base model for this task is a five-layer DNN. Except for APFL, PerFCL and FENDA-FL, all approaches, including locally and centrally trained models, use the same architecture. MOON and FedPer use the first four layers for feature extraction and the final layer as a classification head. In APFL, the same architecture is used but with half of the parameters for each of the global and local components. For FENDA-FL and PerFCL, the global and local modules each have two layers with half the parameters of the full-size model. The classifier module has three layers.

\subsection{Delirium}
A DNN with six layers is used for all non-personalized FL methods. MOON and FedPer use the first five layers as a feature extractor and the final layer as the classifier. APFL’s global and local components are each half the size of the full-sized model. PerFCL uses two five-layer DNNs as the local and global feature extractors and a final linear layer as the classifier. FENDA-FL’s model consists of a two-layer and a four-layer feature extractor as the local and global modules, respectively. The classifier is a linear transform. Note that the FENDA-FL network is asymmetric.

\section{Additional Experimental Details} \label{general_experiment_setup}

For all experiments, clients use an AdamW optimizer \citep{Loshchilov1} with default parameters for local training. The only exception is SCAFFOLD, which explicitly assumes an unmodified learning rate is applied and, thus, uses standard SGD. The client learning rate is tuned for all approaches. For more information on the hyper-parameters tuned for each method, along the ranges tested, best values, and selection methodology, see Appendices \ref{flamby_data_hps} and \ref{gemini_data_hps}. For FedAdam, $\beta_1 = 0.9$, $\beta_2 = 0.99$, and $\tau = 1\mathrm{e}{\text{-}9}$ for all datasets. Finally, all APFL training initializes $\alpha=0.5$. In selecting the FL algorithms for the experiments, FedAdam is chosen as representative of the FedOpt family \citep{McMahan3} because it performed well in the original FLamby experiments. Similarly, the Cyclic method is not considered here as it performed poorly there.

\begin{table*}[ht!]
    \caption{Hyper-parameters tuned for each FL method and FLamby dataset pair. For the ``Range'' column, the sets progress in orders of magnitude. For example, $0.1$ to $1\mathrm{e}{\text{-}4}$ is the set of parameters \{0.1, 0.01, $1\mathrm{e}{\text{-}3}$ $1\mathrm{e}{\text{-}4}$\}. The ``Best'' hyper-parameter values are chosen based on the average aggregated validation loss during federated training over five different training runs.}
    \resizebox{0.99\linewidth}{!}{\begin{tabular}{cccccccc}
        \toprule
        \multicolumn{2}{r}{Task} & \multicolumn{2}{c}{\hspace{2ex}Fed-Heart-Disease} & \multicolumn{2}{c}{Fed-IXI} & \multicolumn{2}{c}{\hspace{1ex}Fed-ISIC2019} \\
        Method & Parameters & Range & Best & Range & Best & Range & Best \\
        \midrule
        FedAvg & Client LR & $0.1$ to $1\mathrm{e}{\text{-}5}$ & $0.1$ & $0.1$ to $1\mathrm{e}{\text{-}5}$ & $1\mathrm{e}{\text{-}3}$ & $0.1$ to $1\mathrm{e}{\text{-}5}$ & $1\mathrm{e}{\text{-}3}$ \\
        \midrule
        \multirow{2}{*}{FedAdam} & Client LR & $0.1$ to $1\mathrm{e}{\text{-}5}$ & $1\mathrm{e}{\text{-}5}$ & $0.1$ to $1\mathrm{e}{\text{-}5}$ & $1\mathrm{e}{\text{-}3}$ & $0.1$ to $1\mathrm{e}{\text{-}5}$ & $1\mathrm{e}{\text{-}5}$ \\
         & Server LR & $0.1$ to $1\mathrm{e}{\text{-}5}$ & $0.1$ & $1.0$ to $1\mathrm{e}{\text{-}4}$ & $0.01$ & $1.0$ to $1\mathrm{e}{\text{-}4}$ & $1\mathrm{e}{\text{-}3}$ \\
        \midrule
        \multirow{2}{*}{FedProx} & Client LR & $1\mathrm{e}{\text{-}3}$ to $1\mathrm{e}{\text{-}4}$ & $1\mathrm{e}{\text{-}3}$ & $0.1$ to $1\mathrm{e}{\text{-}5}$ & $1\mathrm{e}{\text{-}3}$ & $0.1$ to $1\mathrm{e}{\text{-}5}$ & $1\mathrm{e}{\text{-}3}$ \\
         & $\mu$ & $1.0$ to $0.01$ & $0.01$ & $1.0$ to $1\mathrm{e}{\text{-}3}$ & $1\mathrm{e}{\text{-}3}$ & $1.0$ to $1\mathrm{e}{\text{-}3}$ & $0.1$ \\
        \midrule
        \multirow{2}{*}{SCAFFOLD} & Client LR & $0.1$ to $1\mathrm{e}{\text{-}5}$ & $0.1$ & $0.1$ to $1\mathrm{e}{\text{-}5}$ & $0.1$ & $0.1$ to $1\mathrm{e}{\text{-}5}$ & $0.01$ \\
         & Server LR & $0.1$ to $1\mathrm{e}{\text{-}5}$ & $0.1$ & $1.0$ to $1\mathrm{e}{\text{-}4}$ & $1.0$ & $1.0$ to $1\mathrm{e}{\text{-}4}$ & $1.0$ \\
        \midrule
        \multirow{2}{*}{MOON} & Client LR & $0.1$ to $1\mathrm{e}{\text{-}5}$ & $1\mathrm{e}{\text{-}3}$ & $0.1$ to $1\mathrm{e}{\text{-}5}$ & $1\mathrm{e}{\text{-}3}$ & $0.1$ to $1\mathrm{e}{\text{-}5}$ & $1\mathrm{e}{\text{-}3}$ \\
        & $\mu$ & $10$ to $1\mathrm{e}{\text{-}3}$, $5$ & $1\mathrm{e}{\text{-}3}$ & $10$ to $1\mathrm{e}{\text{-}3}$, $5$ & $1\mathrm{e}{\text{-}3}$ & $10$ to $1\mathrm{e}{\text{-}3}$, $5$ & $5$ \\
        \midrule
        \multirow{1}{*}{FedPer} & Client LR & $0.1$ to $1\mathrm{e}{\text{-}5}$ & $1\mathrm{e}{\text{-}3}$ & $0.1$ to $1\mathrm{e}{\text{-}5}$ & $1\mathrm{e}{\text{-}3}$ & $0.1$ to $1\mathrm{e}{\text{-}5}$ & $1\mathrm{e}{\text{-}3}$ \\
        \midrule
        \multirow{2}{*}{Ditto} & Client LR & $0.1$ to $1\mathrm{e}{\text{-}5}$ & $0.1$ & $0.1$ to $1\mathrm{e}{\text{-}5}$ & $1\mathrm{e}{\text{-}3}$ & $0.1$ to $1\mathrm{e}{\text{-}5}$ & $1\mathrm{e}{\text{-}3}$ \\
         & $\lambda$ & $1.0$ to $0.01$ & $0.01$ & $1.0$ to $0.01$ & $1.0$ & $1.0$ to $0.01$ & $1.0$ \\
        \midrule
        \multirow{2}{*}{APFL} & Client LR & $0.1$ to $1\mathrm{e}{\text{-}5}$ & $0.1$ & $0.1$ to $1\mathrm{e}{\text{-}5}$ & $1\mathrm{e}{\text{-}3}$ & $0.1$ to $1\mathrm{e}{\text{-}5}$ & $1\mathrm{e}{\text{-}4}$ \\
         & $\alpha$ LR & $1.0$ to $1\mathrm{e}{\text{-}4}$ & $0.1$ & $1.0$ to $1\mathrm{e}{\text{-}4}$ & $1.0$ & $1.0$ to $1\mathrm{e}{\text{-}4}$ & $0.01$ \\
        \midrule
        \multirow{3}{*}{PerFCL} & Client LR & $0.1$ to $1\mathrm{e}{\text{-}3}$ & $1\mathrm{e}{\text{-}3}$ & $0.1$ to $1\mathrm{e}{\text{-}3}$ & $1\mathrm{e}{\text{-}3}$ & $0.1$ to $1\mathrm{e}{\text{-}3}$ & $1\mathrm{e}{\text{-}3}$ \\
         & $\mu$ & $1.0$ to $0.01$ & $0.1$ & $1.0$ to $0.01$ & $0.1$ & $1.0$ to $0.01$ & $0.01$ \\
         & $\gamma$ & $\{1, 5, 10\}$ & $1.0$ & $\{1, 5, 10\}$ & $10$ & $\{1, 5, 10\}$ & $10$ \\
        \midrule
        \multirow{1}{*}{FENDA-FL} & Client LR & $0.1$ to $1\mathrm{e}{\text{-}5}$ & $1\mathrm{e}{\text{-}3}$ & $0.1$ to $1\mathrm{e}{\text{-}5}$ & $1\mathrm{e}{\text{-}3}$ & $0.1$ to $1\mathrm{e}{\text{-}5}$ & $1\mathrm{e}{\text{-}3}$ \\
        \bottomrule
    \end{tabular}}
    \label{flamby_dataset_hp_studies}
\end{table*}

\section{FLamby Details and Hyper-parameters} \label{flamby_data_hps}

In this section details are provided for the datasets selected from the FLamby benchmark. For more information on items such as preprocessing and feature engineering, see \citep{Terrail1}. The library provided with the benchmark is used to replicate the training parameters, loss functions, model architectures, and other experimental components for consistency. Table \ref{flamby_dataset_hp_studies} details the hyper-parameter sweeps and best values used for the results reported in Section \ref{results}, which differ from the original FLamby settings. Note that the searched values for MOON, PerFCL, and Ditto are informed by the original papers. The best hyper-parameters are chosen by computing the lowest aggregated validation loss seen by the server during FL training, averaged over five training runs. The aggregated loss is computed by averaging each client's validation loss, weighted by the number of examples held by each client. For central and local training, a setup identical to the FLamby benchmark is used. As a result, the best performing parameters optimized in the benchmark are applied for each task. Client dataset sizes for each task are shown in Table \ref{fl_dataset_distributions}.

\begin{table*}[ht!]
    \centering
    \caption{Train and test set sizes across clients in the FLamby datasets, reproduced from \citep{Terrail1}, and GEMINI datasets. Each GEMINI client represents a separate Canadian institution.}
    \resizebox{0.51\linewidth}{!}{\begin{tabular}[t]{ccccc}
    \toprule
    Dataset & Number & Train Size & Test Size \\
    \toprule
    & 0 & 199 & 104 \\
    \cmidrule{2-4}
    Fed-Heart- & 1 & 172 & 89 \\
    \cmidrule{2-4}
    Disease & 2 & 30 & 16 \\
    \cmidrule{2-4}
    & 3 & 85 & 45 \\
    \midrule 
    \multirow{3}{*}[-0.65em]{Fed-IXI} & 0 & 262 & 66 \\
    \cmidrule{2-4}
    & 1 & 142 & 36 \\
    \cmidrule{2-4}
    & 2 & 54 & 14 \\
    \midrule 
    \multirow{6}{*}[-2em]{Fed-ISIC2019} & 0 & 9930 & 2483 \\
    \cmidrule{2-4}
    & 1 & 3163 & 791 \\
    \cmidrule{2-4}
    & 2 & 2691 & 672 \\
    \cmidrule{2-4}
    & 3 & 1807 & 452 \\
    \cmidrule{2-4}
    & 4 & 655 & 164 \\
    \cmidrule{2-4}
    & 5 & 351 & 88 \\
    \bottomrule
    \end{tabular}}
    \hfill
    \resizebox{0.45\linewidth}{!}{\begin{tabular}[t]{cccc}
    \toprule
    Dataset & Number  & Train Size & Test Size \\
    \toprule
    \multirow{4}{1.4cm}[-2.5em]{GEMINI Mortality} & 0 &  13234 & 3308 \\
    \cmidrule{2-4}
    & 1 &  21018 & 5254 \\
    \cmidrule{2-4}
    & 2 &  14180 & 3545 \\
    \cmidrule{2-4}
    & 3 &  15251 & 3813 \\
    \cmidrule{2-4}
    & 4 &  14441 & 3610 \\
    \cmidrule{2-4}
    & 5 &  12718 & 3180 \\
    \cmidrule{2-4}
    & 6 &  23820 & 5955 \\
    \midrule
    \multirow{3}{1.3cm}[-3em]{GEMINI Delirium} & 0 &  660 & 101 \\
    \cmidrule{2-4}
    & 1 &  590 & 99 \\
    \cmidrule{2-4}
    & 2 & 1218 & 199 \\
    \cmidrule{2-4}
    & 3 & 494 & 76 \\
    \cmidrule{2-4}
    & 4 & 574 & 109 \\
    \cmidrule{2-4}
    & 5 & 1392 & 258 \\
    \bottomrule
    \end{tabular}}
    \label{fl_dataset_distributions}
\end{table*}

\paragraph{Fed-Heart-Disease:}

The Fed-Heart-Disease dataset consists of four clients. The loss function is binary cross-entropy and the metric is standard accuracy. A batch size of $4$ is used for all training. Central and local training uses a learning rate of $0.001$ over $50$ epochs with an AdamW optimizer.

\paragraph{Fed-IXI:}

The Fed-IXI dataset incorporates data from three distinct centres. Two different scanning systems produced the 3D images in the dataset. The brain scans are labeled with binary segmentation masks. For training, a Dice loss \citep{Dice1} is used with an $\epsilon=10^{-9}$. The metric is the Dice score between the predicted and labeled segmentations. A constant batch size of $2$ is used for all training. Both central and local training apply a learning rate of $0.001$ and an AdamW optimizer for $10$ epochs.

\paragraph{Fed-ISIC2019:}

The Fed-ISIC2019 dataset is split across six clients. Three of the client datasets are derived from the same source, the Medical University of Vienna, but were generated by different imaging devices. A weighted focal loss function \citep{Lin1} is used to fine-tune EfficientNet-B0 architectures and balanced accuracy is measured to assess performance. All training runs use a batch size of $64$. As seen in Table \ref{fl_dataset_distributions}, the number of data points held by each client varies dramatically.

\subsection{Sub-Optimal Hyper-Parameters For Checkpointing Study} 
\label{sub_optimal_fed_ixi}

For all strategies, except PerFCL, a client-side learning rate of $0.1$ is applied. For FedAdam and SCAFFOLD, a server-side learning rate of $0.1$ and $1.0$, respectively, is used. FedProx sets $\mu$ = $1\mathrm{e}{\text{-}3}$, APFL applies a learning rate for $\alpha$ of 1.0, MOON sets $\mu$ to $1\mathrm{e}{\text{-}3}$, and Ditto uses $\lambda = 0.01$. Finally, PerFCL uses a client-side learning rate of $1\mathrm{e}{\text{-}3}$, $\mu = 0.01$, and $\gamma = 1$. 

\begin{table*}[ht!]
    \centering
    \caption{Hyper-parameters tuned for each FL method, along with local and central training, across GEMINI tasks. For the ``Range'' column, the sets progress in orders of magnitude. For example, $0.01$ to $1\mathrm{e}{\text{-}4}$ is the set of parameters \{0.01, $1\mathrm{e}{\text{-}3}$ $1\mathrm{e}{\text{-}4}$\}. The ``Best'' hyper-parameter values are chosen based on the average aggregated validation loss during federated training over five different training runs.}
    \resizebox{\linewidth}{!}{\begin{tabular}[t]{cccccc}
        \toprule
        \multicolumn{2}{r}{Task} & \multicolumn{2}{c}{\hspace{3ex}Mortality} & \multicolumn{2}{c}{Delirium}\\
        Method & Params. & Range & Best & Range & Best\\
        \midrule
        FedAvg & Client LR & $0.01$ to $1\mathrm{e}{\text{-}4}$ & $1\mathrm{e}{\text{-}3}$ & $0.01$ to $1\mathrm{e}{\text{-}4}$ & $1\mathrm{e}{\text{-}3}$ \\
        \midrule
        \multirow{2}{*}{FedAdam} & Client LR & $0.01$ to $1\mathrm{e}{\text{-}4}$ & $1\mathrm{e}{\text{-}4}$ & $0.01$ to $1\mathrm{e}{\text{-}4}$ & $1\mathrm{e}{\text{-}4}$ \\
         & Server LR & $1.0$ to $1\mathrm{e}{\text{-}3}$ & $1\mathrm{e}{\text{-}3}$ & $1.0$ to $1\mathrm{e}{\text{-}3}$ & $1\mathrm{e}{\text{-}3}$\\
        \midrule
        \multirow{2}{*}{FedProx} & Client LR & $1.0$ to $1\mathrm{e}{\text{-}4}$ & $0.01$ & $0.1$ to $1\mathrm{e}{\text{-}4}$ & $1\mathrm{e}{\text{-}3}$ \\
         & $\mu$ & $1.0$ to $0.01$  & $0.1$ & $1.0$ to $0.01$ & $0.1$ \\
        \midrule        
        \multirow{2}{*}{SCAFFOLD} & Client LR & $0.01$ to $1\mathrm{e}{\text{-}4}$ & $0.01$ & $0.01$ to $1\mathrm{e}{\text{-}4}$ & $1\mathrm{e}{\text{-}3}$ \\
         & Server LR & $1.0$ to $1\mathrm{e}{\text{-}3}$ & $1.0$ & $1.0$ to $1\mathrm{e}{\text{-}4}$ & $1\mathrm{e}{\text{-}3}$\\
         \midrule        
        \multirow{2}{*}{MOON} & Client LR & $0.1$ to $1\mathrm{e}{\text{-}5}$ & $1\mathrm{e}{\text{-}4}$ & $0.1$ to $1\mathrm{e}{\text{-}5}$ & $1\mathrm{e}{\text{-}4}$ \\
        & $\mu$ & $10$ to $1\mathrm{e}{\text{-}3}$, $5$ & $1\mathrm{e}{\text{-}3}$ & $10$ to $1\mathrm{e}{\text{-}3}$, $5$ & $1.0$ \\
        \midrule       
        \multirow{1}{*}{FedPer} & Client LR & $0.1$ to $1\mathrm{e}{\text{-}5}$ & $1\mathrm{e}{\text{-}4}$ & $0.1$ to $1\mathrm{e}{\text{-}5}$ & $1\mathrm{e}{\text{-}5}$ \\
        \midrule
        \multirow{2}{*}{Ditto} & Client LR & $0.1$ to $1\mathrm{e}{\text{-}5}$ & $1\mathrm{e}{\text{-}4}$ & $0.1$ to $1\mathrm{e}{\text{-}5}$ & $1\mathrm{e}{\text{-}5}$ \\
         & $\lambda$ & $1.0$ to $0.01$ & $0.01$ & $1.0$ to $0.01$ & $0.01$ \\
        \midrule
        \multirow{2}{*}{APFL} & Client LR & $0.1$ to $1\mathrm{e}{\text{-}4}$ & $0.01$ & $0.01$ to $1\mathrm{e}{\text{-}5}$ & $1\mathrm{e}{\text{-}4}$ \\
         & $\alpha$ LR & $1.0$ to $1\mathrm{e}{\text{-}3}$ & $0.01$ & $1.0$ to $1\mathrm{e}{\text{-}3}$ & $0.01$ \\
        \midrule        
        \multirow{3}{*}{PerFCL} & Client LR & $0.1$ to $1\mathrm{e}{\text{-}3}$ & $1\mathrm{e}{\text{-}3}$ & $0.1$ to $1\mathrm{e}{\text{-}3}$ & $1\mathrm{e}{\text{-}3}$ \\
        & $\mu$ & $1.0$ to $0.01$ & $0.1$ & $1.0$ to $0.01$ & $1.0$ \\
        & $\gamma$ & $\{1, 5, 10\}$ & $10$ & $\{1, 5, 10\}$ & $1.0$ \\
        \midrule
        \multirow{1}{*}{FENDA-FL} & Client LR & $0.01$ to $1\mathrm{e}{\text{-}5}$ & $1\mathrm{e}{\text{-}4}$ & $0.01$ to $1\mathrm{e}{\text{-}5}$ & $1\mathrm{e}{\text{-}4}$ \\ 
        \bottomrule
    \end{tabular}
    \begin{tabular}[t]{cccccc}
        \toprule
        \multicolumn{2}{r}{Task} & \multicolumn{2}{c}{\hspace{3ex}Mortality} & \multicolumn{2}{c}{Delirium}\\
        Method & Params. & Range & Best & Range & Best\\
        \midrule
        \multirow{1}{*}{Local 0} & LR & $0.01$ to $1\mathrm{e}{\text{-}4}$ & $0.01$ & $0.01$ to $1\mathrm{e}{\text{-}4}$ & $0.01$ \\
        \midrule
        \multirow{1}{*}{Local 1} & LR & $0.01$ to $1\mathrm{e}{\text{-}4}$ & $1\mathrm{e}{\text{-}3}$ & $0.01$ to $1\mathrm{e}{\text{-}4}$ & $0.01$ \\
        \midrule
        \multirow{1}{*}{Local 2} & LR & $0.01$ to $1\mathrm{e}{\text{-}4}$ & $0.01$ & $0.01$ to $1\mathrm{e}{\text{-}4}$ & $1\mathrm{e}{\text{-}4}$ \\
        \midrule
        \multirow{1}{*}{Local 3} & LR & $0.01$ to $1\mathrm{e}{\text{-}4}$ & $0.01$ & $0.01$ to $1\mathrm{e}{\text{-}4}$ & $1\mathrm{e}{\text{-}3}$ \\
        \midrule
        \multirow{1}{*}{Local 4} & LR & $0.01$ to $1\mathrm{e}{\text{-}4}$ & $0.01$ & $0.01$ to $1\mathrm{e}{\text{-}4}$ & $1\mathrm{e}{\text{-}3}$ \\
        \midrule
        \multirow{1}{*}{Local 5} & LR & 0.01 to $1\mathrm{e}{\text{-}4}$ & $0.01$ & $0.01$ to $1\mathrm{e}{\text{-}4}$ & $1\mathrm{e}{\text{-}3}$ \\
         \midrule
        \multirow{1}{*}{Local 6} & LR & $0.01$ to $1\mathrm{e}{\text{-}4}$ & $0.01$ & $-$  & $-$  \\ 
        \midrule
        \multirow{1}{*}{Central} & LR & $0.01$ to $1\mathrm{e}{\text{-}5}$ & $1\mathrm{e}{\text{-}4}$ & $0.01$ to $1\mathrm{e}{\text{-}5}$  & $1\mathrm{e}{\text{-}4}$  \\ 
        \bottomrule
    \end{tabular}}
    \label{gemini_dataset_hp_studies}
\end{table*}

\section{GEMINI Details and Hyper-parameters} \label{gemini_data_hps}

GEMINI is a research network in which hospitals share data with a central repository. As the repository contains detailed patient information, access to GEMINI data is granted only after approval, in accordance with the network's data sharing agreement and research ethics protocols. As in the FLamby experiments, hyper-parameters are chosen based on the minimum weighted-average validation loss computed by the server. The sweep space and best hyper-parameters for each method are reported in Table \ref{gemini_dataset_hp_studies}. Hyper-parameter sweeps are also performed for central and local training for the GEMINI tasks, the results of which are also found in the table. In addition, Table \ref{fl_dataset_distributions} displays the client data distributions. An overview of the data preparation steps is provided in subsequent sections. The code is available to be shared with researchers upon receiving access to GEMINI.

\paragraph{Federated Pre-Processing}
Data alignment in FL pipelines is challenging. For example, categorical features require special attention in a federated setting. For the GEMINI tasks, a federated pre-processing step is included for such features to synchronize vectorization and ensure that the resulting feature spaces remain aligned across clients. For clients to consistently map such features to a numerical space, we assume that the server has oracle knowledge of all possible feature values and provides one-hot mappings to all clients. 

\subsection{Mortality Prediction} \label{gemini_mortality_details}

For this task, there are seven hospitals, each representing a client with its own private data. A binary cross-entropy loss is used for training, with a batch size of $64$. The number of samples for mortality prediction is large compared to other tasks. A careful analysis resulted in the selection of $35$ features that were subsequently processed by a standard data pipeline, which included data vectorization and scaling. For more details about the data pre-processing, see \citep{Subasri2023}.  A non-exhaustive list of features includes: readmission status, whether the patient is transferred from an acute care institution, length of stay (LOS) in days, CD-10-CA classification code, counts of previous encounters, diagnosis type, and age. Patient characteristics are described in Table \ref{hyper-mortality_statics} for a subset of features.

\begin{table}[ht!]
    \centering
    \caption{Feature characteristics for the Mortality prediction dataset conditioned on label.}
    \begin{tabular}{lccccccc}
    \toprule
        Mortality Label & \multicolumn{7}{c}{False} \\
        \midrule
         Client & 0 & 1 & 2 & 3 & 4 & 5 & 6 \\
        \midrule
        Patients & 15122 & 22719 & 16127 & 16645 & 16250 & 14083 & 27448 \\
        Avg. Prev. Encounters &  0.68 & 0.59 & 1.19 & 0.72 & 0.82 &  0.89 & 0.59 \\
        Avg. LOS (Days) & 8.71 & 10.08 & 7.89 & 8.46 & 7.54 & 9.02 & 8.43 \\
        \% Emergent Triage &  60.55 & 46.83 &  50.88 & 34.71 & 50.01 & 42.7 &  46.11\\
        \% Nursing Home & 10.5 & 11.83 & 4.19 & 6.41 & 4.48 & 10.23 & 8.29 \\
        \midrule
        Mortality Label & \multicolumn{7}{c}{True} \\
         \midrule
         Client & 0 & 1 & 2 & 3 & 4 & 5 & 6 \\
         \midrule
        Patients & 1420 & 3553 & 1598 & 2419 & 1801 & 1815 & 2327\\
        Avg. Prev. Encounters & 1.36 & 1.16 &  1.53 & 1.28 & 1.31 & 1.62 & 1.11\\
        Avg. LOS (Days) & 18.96 & 17.66 & 17.58 & 14.52 & 12.74 & 14.06 & 17.03\\
        \% Emergent Triage & 57.81 & 47.62 & 55.13 & 44.85 & 59.57 &  54.15 & 49.46\\
        \% Nursing Home & 30.77 & 29.63 & 13.82 & 12.48 & 8.21 & 22.09 & 17.61 \\
        \bottomrule
    \end{tabular}
    \label{hyper-mortality_statics}
\end{table}

\subsection{Delirium Prediction}  \label{gemini_delirium_details}

For this task, data is drawn from six hospitals, each participating in FL as a distinct client. As seen in Table \ref{fl_dataset_distributions}, the number of samples in this task is smaller than in mortality prediction. This is because delirium labels are provided through a rigorous and time-intensive process of medical record review by a trained expert \citep{Inouye1}. This is one of the only validated methods for such labeling, as it is poorly captured in routine administrative or clinical data. A wide variety of medical features is used, such as count of room transfer and ICU transfer, whether the patient has entered an intensive care unit since their admission, length of stay in days, laboratory-based Acute Physiology Score, ICU length of stay in hours, duration of the patient’s stay in the ER in hours, whether admission was via ambulance, triage level, total number of transfusions, type of blood product or component transfused, medication names, text description of radiology images, among many others. The feature engineering pipeline results in $1,119$ features. 

The data pre-processing steps are the following. One-hot encoding is used for categorical features such as gender, discharge disposition, and admission category. A TF-IDF vectorizer is used to vectorize text-based columns such as radiology image and medication descriptions. A \texttt{MaxAbsScaler} is used for data scaling. Finally, during training, resampling is performed to balance negative and positive samples and increase the dataset size. Resampling is done separately in each client based on the client’s data. Note that resampling is not performed on the test datasets. Similar to the mortality task, we assume oracle knowledge in the server, enabling clients to map categorical features into a shared space. A batch size of $64$ is used for all methods. In contrast to the Mortality dataset, the patient population for Delirium is fairly small. Therefore, no further statistics or patient characteristics are provided to prevent possible leakage of private information.

\begin{table*}[ht!]
    \centering
    \caption{Results corresponding to different FENDA-FL architecture variations on the Delirium task. ``Layers (G, L)'' and ``Latent (G, L)'' refer to the number of dense layers assigned to the global (G) and local (L) feature extractors, respectively, along with the size of the corresponding latent spaces.}
    \resizebox{0.85\linewidth}{!}{\begin{tabular}{ccccc}
    \toprule
    Layers (G, L) & Latent (G, L) & Parameters & Acc. & AUC/ROC \\
    \midrule
    $4$, $4$ & $64$, $64$ & $8,631,424$ & $0.7926$ ($0.0071$) & $0.8222$ ($0.0036$) \\
    \midrule
    $4$, $2$ & $64$, $64$ & $ 8,729,600$ & $0.8024$ ($0.0037$) & $0.8368$ ($0.0039$) \\
    \midrule
    $4$, $2$ & $128$, $8$ & $8,717,448$ & $\mathbf{0.8064}$ ($0.0036$) & $\mathbf{0.8518}$ ($0.0017$) \\
    \midrule
    $2$, $4$ & $64$, $64$ & $8,729,600$ & $0.7935$ ($0.0073$) & $0.8153$ ($0.0059$) \\
    \midrule
    $2$, $4$ & $8$, $128$ & $8,717,448$ & $0.7935$ ($0.0073$) & $0.8153$ ($0.0059$) \\
    \bottomrule
    \end{tabular}}
    \label{delirium_architecture_perturbations}
\end{table*}

\section{FENDA-FL Architecture Studies}
\label{architecture_ablation}

As discussed in Section \ref{fenda_introduction}, an important advantage of FENDA-FL is that the local and global feature extraction modules need not be symmetric with equal latent space sizes. This allows for the injection of inductive bias through architecture choice. For example, in situations where global models perform well, one may assign more parameters and a larger latent space to FENDA-FL's global feature extractor. In this section, we show that such choices can materially boost performance. 

\begin{table*}[ht!]
    \centering
    \caption{Results corresponding to different FENDA-FL architecture variations on the Fed-IXI task. ``Depth (G, L)'' and ``Latent (G, L)'' refer to the number of projection layers used in the global (G) and local (L) U-Net feature extractors, respectively, along with the size of the latent space for each voxel.}
    \resizebox{0.85\linewidth}{!}{\begin{tabular}{ccccc}
    \toprule
    Strategy & Depth (G, L) & Latent (G, L) & Parameters & Dice \\
    \midrule
    \multirow{2}{*}{APFL} & $3$, $3$ & $11$, $11$ & $1, 859, 928$ & $0.9855$ ($0.0001$) \\
    \cmidrule{2-5}
    & $3$, $3$ & $8$, $8$ & $984, 624 $ & $0.9864$ ($0.0002$) \\
    \midrule
    \multirow{5}{*}{FENDA-FL} & $3$, $3$ & $11$, $11$ & $1, 859, 924$ & $0.9859$ ($0.0001$) \\
    \cmidrule{2-5}
     & $3$, $2$ & $14$, $14$ & $1, 824, 740$ & $0.9864$ ($0.0000$)\\
    \cmidrule{2-5}
     & $2$, $3$ & $14$, $14$ & $1, 824, 740$ & $\mathbf{0.9865}$ ($0.0000$) \\
    \cmidrule{2-5}
     & $3$, $2$ & $16$, $8$ & $2, 070, 692$ & $0.9859$ ($0.0000$) \\
    \cmidrule{2-5}
     & $2$, $3$ & $8$, $16$ & $2, 070, 692$ & $0.9860$ ($0.0000$) \\
    \bottomrule
    \end{tabular}}
    \label{fedixi_architecture_perturbations}
\end{table*}

\begin{figure*}[ht!]
    \centering
    \includegraphics[scale=0.16]{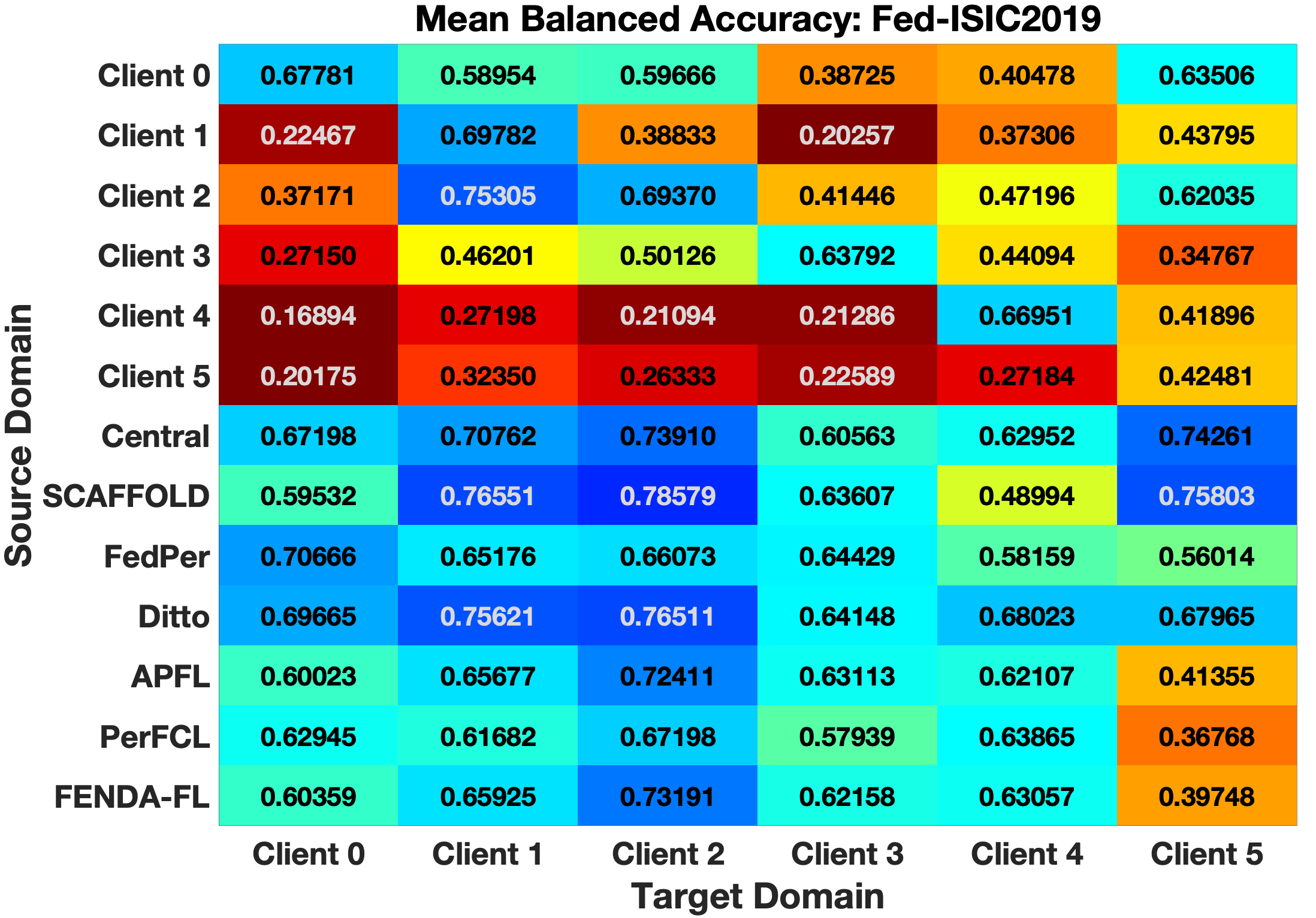}
    \caption{Generalization measurements for a selection of FL and non-FL approaches. Rows correspond to a model trained using either client-only data, centralized data, or with the indicated FL method. Columns denote the average performance of the model on the local test data of the named client.}
    \label{fed_isic_generalization}
\end{figure*}

Results for variations in the FENDA-FL model architecture for the Delirium task are reported in Table \ref{delirium_architecture_perturbations}.
It is evident that assigning more layers and a larger latent space to the global module is beneficial. Given that FedProx, a global approach, is performing this task well, see Table \ref{delirium_results_table}, such an emphasis makes sense. However, the use of an, albeit smaller, personal module allows FENDA-FL to surpass FedProx in terms of AUC/ROC and be competitive with respect to accuracy, despite having fewer parameters.

Table \ref{fedixi_architecture_perturbations} shows results for FENDA-FL architecture perturbations for the Fed-IXI task. While the performance changes are modest, there is an advantage to using a larger local feature extraction module for this dataset. Doing so brings the FENDA-FL model results ahead of the previously best performing strategy, APFL, for this problem. Note that a larger FENDA-FL model is used to allow for latent space imbalance in the U-Net, but that APFL does not benefit from such an increase.

\section{Continued Analysis of Fed-ISIC2019 Results} \label{fed_isic_analysis}

For Fed-ISIC2019, many of the personalized FL approaches, excluding Ditto, produced lower than expected performance, see Table \ref{flamby_results_table}. Some insights can be found in Figure \ref{fed_isic_generalization}. SCAFFOLD, the best non-personalized FL method, outperforms the personalized models on test data drawn from Client 1 and Client 2. This is reversed when considering Client 4. However, the most significant difference is that most personalized models perform quite poorly on the test data of Client 5. Strikingly, Ditto does not suffer from the same precipitous drop, performing well across all client data.

Notably, Client 5 has the smallest dataset by a large margin; see Table \ref{fl_dataset_distributions}. Moreover, the data appears to be of poor quality, with locally trained models only achieving a balanced accuracy of $0.425$. Alternatively, models trained only on data from Client 0 generalize well to Client 5. We hypothesize that the low quality of Client 5's data impedes FL model training for most personalized methods. Further, we posit that SCAFFOLD's control-variate corrections and the strong regularization of Ditto to the global model provide resilience to such local client drift. In future work, we aim to consider ways of overcoming this issue for other methods. Some potential avenues include additional loss terms to ensure the global and local feature extractors remain individually useful for classification or using a global FL approach to initialize model components as a warm start. 

\section{Momentum-Based Server-Side Optimizers and BN} \label{momentum_based_optimizers_and_bn}

Certain layers in deep learning models accumulate state throughout training that must remain strictly positive. Batch normalization (BN) \citep{Ioffe1} is one such layer. During training, the layer accumulates mean and variance estimates of preceding layer activations. These estimates are used to perform normalization during inference. When using FedAvg, these estimates are aggregated, like all other parameters, using weighted averaging. Thus, strictly positive local quantities result in strictly positive aggregated quantities. However, when using momentum-based server-side optimization, as in FedAdam, this property is no longer preserved. This is particularly problematic for BN, where the square root of the variance is computed to perform normalization.

\begin{algorithm2e}
\caption{Simplified FedAdam calculation with a single parameter and constant updates, $\tilde{x} = 0.1$, from clients over $30$ rounds.}
\label{fedadam_algo}
\SetAlgoLined
~\\
$x_t = x_0$, $m_t = 0$, $v_t = 0$
~\\
\For{$t$ \text{from} 1 \text{to} 30}
{
	$\Delta_t = \tilde{x} - x_t$
    ~\\
    $m_t = \beta_1 m_t + (1-\beta_1)\Delta_t$
    ~\\
    $v_t = \beta_2 v_t + (1-\beta_2)\Delta_t^2$
    ~\\
    $x_t = x_t + \eta \frac{m_t}{\sqrt{v_t} + \tau}$
}
\end{algorithm2e}

As an illustration of this phenomenon, consider learning a single parameter $x$ from an initial guess of $x_0 = 2.0$. Set the FedAdam parameters to $\beta_1 = 0.9$, $\beta_2 = 0.9$, $\tau = 1\mathrm{e}{\text{-}9}$, and a learning rate of $\eta=0.1$. Assume that each client calculates a new parameter of $\tilde{x} = 0.1$ at each iteration. Algorithm \ref{fedadam_algo} shows the iterations for this simplified setting. After $30$ steps, $x_t = -0.204$, despite $\tilde{x} > 0$ for every iteration. This drift is due to the momentum associated with the updates. Therefore, in settings where BN is present and the FedAdam strategy is employed, batch tracking is shut off to avoid such issues. It should be noted that this is qualitatively similar to the FedBN strategy in \citep{Li1}.

\section{FENDA-FL: Distribution Shift and Domain Adaptation} \label{formal_fenda_desription}

\subsection{Distribution Shifts in FL}

Distribution shifts arise when the joint distribution of inputs and targets varies between datasets \citep{quinonero2008dataset}. For standard ML, shifts occur between the train and test data, such that $P_{\text{train}}(X, Y) \neq P_{\text{test}}(X, Y)$, leading to poor generalization. The most common shifts include label, covariate, and concept drift \citep{zhang2021dive}. Label shift is characterized by different label distributions, $P(Y)$, but constant class conditioned distributions, $P(X \vert Y)$. Covariate shift occurs when the marginal distribution, $P(X)$, varies  across datasets while the conditional distribution, $P(Y \vert X)$, remains the same. Concept drift arises when the class conditional distribution $P(Y \vert X)$ varies and $P(Y)$ remains fixed. 

In FL, distribution shifts are common among clients, due to variations in the underlying data generating process. These variations may stem, for example, from differences between sensors, geographies and time periods involved in data collection \citep{Kairouz1}. In this regime, FedAvg often exhibits slow convergence and poor performance. Many works have attempted to address this for various specific types of shifts \citep{Karimireddy1, Deng1, li2023federated}. 

As a personalized method, FENDA-FL allows each client to train a custom model while maintaining the benefit of using FL to jointly learn a subset of weights. In particular, FENDA-FL has a locally-optimized feature extractor and classification head, in addition to a global feature extractor learned via FedAvg. The local feature extractor and classification head endow client models with a degree of resilience to label and covariate shift. They may adapt feature representations and predictions to lean more or less heavily on global representations or modify prediction distributions to match local statistics. Beyond such shifts, FENDA-FL is able to handle concept drift where the conditional distribution, $P(Y \vert X)$, varies across clients. For example, in the extreme case, this is accomplished by the local classification head ignoring global features altogether. This is not true of non-personalized FL models, which assume a single function exists that obtains low error across all clients. 

FENDA-FL is also resilient to extreme distribution shifts. As demonstrated in Section \ref{delirium_het_study}, FENDA-FL is able to obtain improved performance compared to locally trained models even when the feature spaces of the clients are not aligned, whereas other benchmark approaches fall short. However, one type of distribution shift that is not admissible in the proposed approach is misaligned label spaces. This is an important setting to explore in future work.

\subsection{Domain Adaptation and Generalization}

FENDA-FL is connected to Domain Adaptation (DA). Generally, DA involves learning a model from source datasets that generalizes to target datasets. During training, subsets of the target datasets, typically unlabeled or sparsely labeled, are available. Federated DA considers the setting of distributed source and target datasets. A server may host the source dataset coupled with client-based, unlabelled datasets \citep{yao2022federated} or vice-versa \citep{peng2019federated}. Alternatively, both source and target datasets may be distributed \citep{Shen1, Song1}. Similarly, Federated Domain Generalization focuses on the task of training a model on distributed, private source domains that generalizes to \emph{unseen} target domains \citep{li2023federated}. Such methods focus on training models for out-of-distribution clients rather than (personalized) models optimized for known client data. These settings are related to those considered here but have different objectives and data constraints. However, FENDA-FL establishes a novel and useful conceptual bridge between FL and DA.

\end{document}